\documentclass[lettersize,journal]{IEEEtran}

\usepackage{array}
\usepackage{amssymb}
\usepackage{graphicx}
\usepackage{cite}
\usepackage{subfigure,color}

\usepackage{multirow,amsmath}

\newcommand{\highlight}[1]{\textcolor{black}{#1}}
\newcommand{\etal}{\textit{et al.}}

\newcommand{\tabincell}[2]{\begin{tabular}{@{}#1@{}}#2\end{tabular}}

\begin{document}

\title{CT-Net: Arbitrary-Shaped Text Detection via Contour Transformer}
%Arbitrary-Shaped Text Detection by Progressive Contour Transformer
%multi-stage contour-based scene text detector with contour transformer

\author{Zhiwen~Shao,
        Yuchen~Su,
        Yong~Zhou,
        Fanrong~Meng,
        Hancheng~Zhu,
        Bing~Liu,
        and~Rui~Yao
% \thanks{Manuscript received December, 2021. (Corresponding authors: Yong~Zhou; Zhiwen~Shao; Fanrong~Meng.)}
\thanks{Manuscript received November, 2022. This work was supported in part by the National Natural Science Foundation of China under Grant 62106268, in part by the China Postdoctoral Science Foundation under Grant 2023M732223, in part by the High-Level Talent Program for Innovation and Entrepreneurship (ShuangChuang Doctor) of Jiangsu Province under Grant JSSCBS20211220, and in part by the Talent Program for Deputy General Manager of Science and Technology of Jiangsu Province under Grant FZ20220440. It was also supported in part by the National Natural Science Foundation of China under Grants 62272461, 62101555, 62276266, and 62172417, in part by the Natural Science Foundation of Jiangsu Province under Grants BK20201346 and BK20210488, and in part by the Shanghai Sailing Program under Grant 23YF1410500. (Zhiwen~Shao and Yuchen~Su contributed equally to this work. Corresponding author: Yong~Zhou.)}%
\thanks{Z. Shao is with the School of Computer Science and Technology, China University of Mining and Technology, Xuzhou 221116, China, also with the Engineering Research Center of Mine Digitization, Ministry of Education of the People’s Republic of China, Xuzhou 221116, China, and also with the Department of Computer Science and Engineering, Shanghai Jiao Tong University, Shanghai 200240, China (e-mail: zhiwen\_shao@cumt.edu.cn).}
\thanks{Y. Su, Y. Zhou, F. Meng, H. Zhu, B. Liu, and R. Yao are with the School of Computer Science and Technology, China University of Mining and Technology, Xuzhou 221116, China, and also with the Engineering Research Center of Mine Digitization, Ministry of Education of the People’s Republic of China, Xuzhou 221116, China (e-mail: \{yuchen\_su; yzhou; mengfr; zhuhancheng; liubing; ruiyao\}@cumt.edu.cn).}
}

\markboth{IEEE Transactions on Circuits and Systems for Video Technology,~Vol.~X, No.~X, X}%
{Shell \MakeLowercase{\textit{et al.}}: Bare Demo of IEEEtran.cls for IEEE Journals}

\maketitle

\begin{abstract}
Contour based scene text detection methods have rapidly developed recently, but still suffer from inaccurate front-end contour initialization, multi-stage error accumulation, or deficient local information aggregation. To tackle these limitations, we propose a novel arbitrary-shaped scene text detection framework named CT-Net by progressive contour regression with contour transformers. Specifically, we first employ a contour initialization module that generates coarse text contours without any post-processing. Then, we adopt contour refinement modules to adaptively refine text contours in an iterative manner, which are beneficial for context information capturing and progressive global contour deformation. Besides, we propose an adaptive training strategy to enable the contour transformers to learn more potential deformation paths, and introduce a re-score mechanism that can effectively suppress false positives. Extensive experiments are conducted on four challenging datasets, which demonstrate the accuracy and efficiency of our CT-Net over state-of-the-art methods. Particularly, CT-Net achieves F-measure of $86.1$ at $11.2$ frames per second (FPS) and F-measure of $87.8$ at $10.1$ FPS for CTW1500 and Total-Text datasets, respectively.
\end{abstract}


\IEEEpeerreviewmaketitle

\section{Introduction}
\IEEEPARstart{O}{ver} the past few years, scene text detection has attracted ever-increasing interests in the computer vision community, due to its ubiquitous applications in many fields, such as product search \cite{xiong2016text}, scene understanding \cite{yi2014scene}, and autonomous driving \cite{hong2019textplace}. Benefited from the rapid development of deep learning\highlight{~\cite{xu2022x,shu2022multi}}, scene text detection methods have achieved remarkable progress \cite{cheng2019direct,wang2019efficient,2020Learning,wang2020contournet,xing2021boundary}. However, due to the complex backgrounds and large variance of scene texts in color, size, texture, etc., arbitrary-shaped scene text detection remains a challenge.

To explore an arbitrary-shaped scene text detector, segmentation based and connected component based approaches represent text instances with pixel-level classification masks or sequential components from a local modeling perspective. Although the shapes of the scene texts are easy to model,%does not constrain these approaches,
they suffer from the lack of noise resistance and complicated heuristic post-processing. More importantly, they commonly lack attention to the global geometric distribution that reflects the overall layouts of the text contours.
%the bottom-up methods have sparked a new wave recently, which locate text regions based on pixel-level predictions, while predicting auxiliary information for each pixel and clustering text pixels into different instances. However, these methods are sensitive to noise, rely on heuristic post-processing, and lack global awareness of text boundaries.

To obtain the global geometric distribution, some single-stage contour based methods are proposed to model the text contours directly. For example, ABCNet \cite{liu2020abcnet} and FCENet \cite{zhu2021fourier} represent text contours with Bessel points and Fourier coefficients, respectively. However, these methods regress the text contours based on the limited features at the positive point and perceive the texts with complex geometric layouts only in single stage, resulting in the loss of the predicted contour details. Therefore, some multi-stage contour based methods are proposed to progressively locate the texts, % by multiple perceptions on the features of contour vertexes,
%to progressively regress the text contours,
which first generate the initial contours by a detector and then progressively refine the initial contours by multiple contour refinement modules.
%to resolving the problem of inaccurate perception once.
Although the existing multi-stage contour based methods have achieved excellent detection performance, they still have three main problems to be resolved.

\begin{figure}[t]
    \centering
    \includegraphics[width=\linewidth]{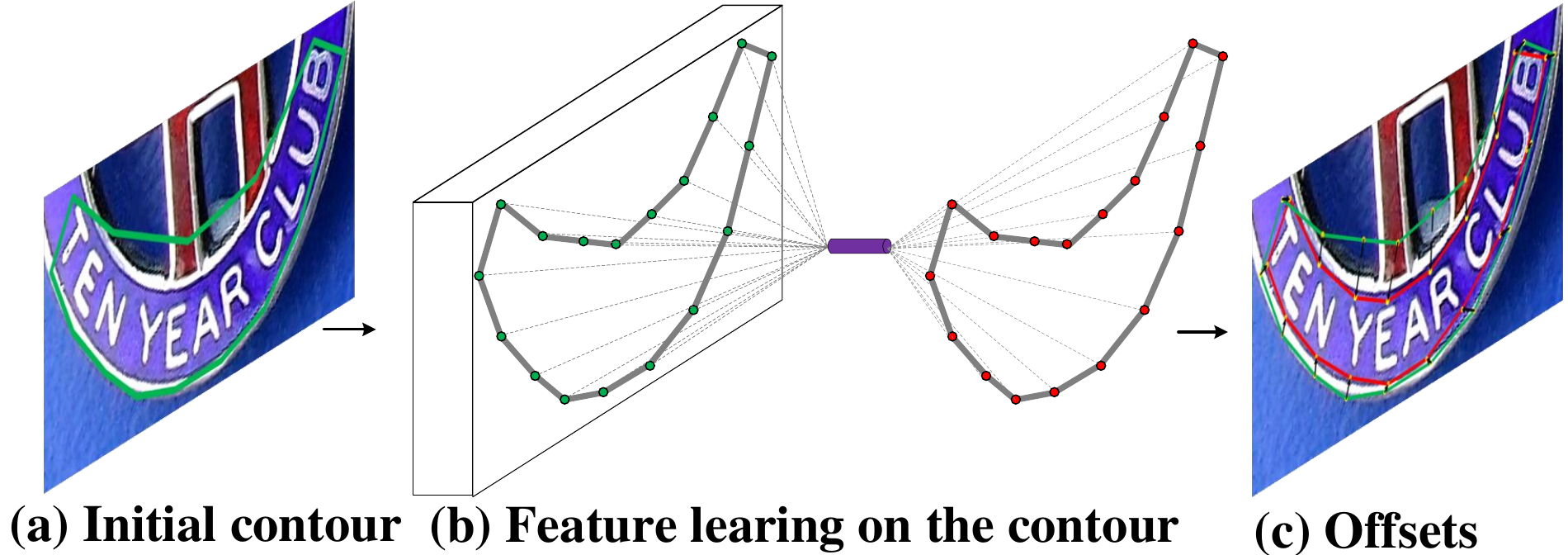}
    \caption{\textbf{The overall idea of our CT-Net.} (a) Generation of initial contour with our contour  initialization module; (b) Vertex-wise feature learning, in which the green and the red nodes denote the input features and the output results, respectively, and the purple cylinder denotes our contour transformer; (c) Offsets are regressed at each vertex to deform the contour to the text boundary.}
    \label{fig-basic}
\end{figure}

\highlight{\textit{First, inaccurate front-end contour initialization restricts the performance.}} For example, PCR \cite{dai2021progressive} regresses directly from oriented text proposals to arbitrary-shaped contours. This large gap between the oriented text proposals and the ground-truth text contours may lead to many unreasonable deformation paths and huge training difficulty, especially for highly-curved texts. TextBPN \cite{zhang2021adaptive} uses a segmentation based approach to generate initial contours. However, this bottom-up approach cannot take advantage of the prior knowledge that there are no holes inside the texts, maybe resulting in a text instance split into multiple ones, especially for large character spacing texts.

\begin{figure*}
\centering
\includegraphics[width=\linewidth]{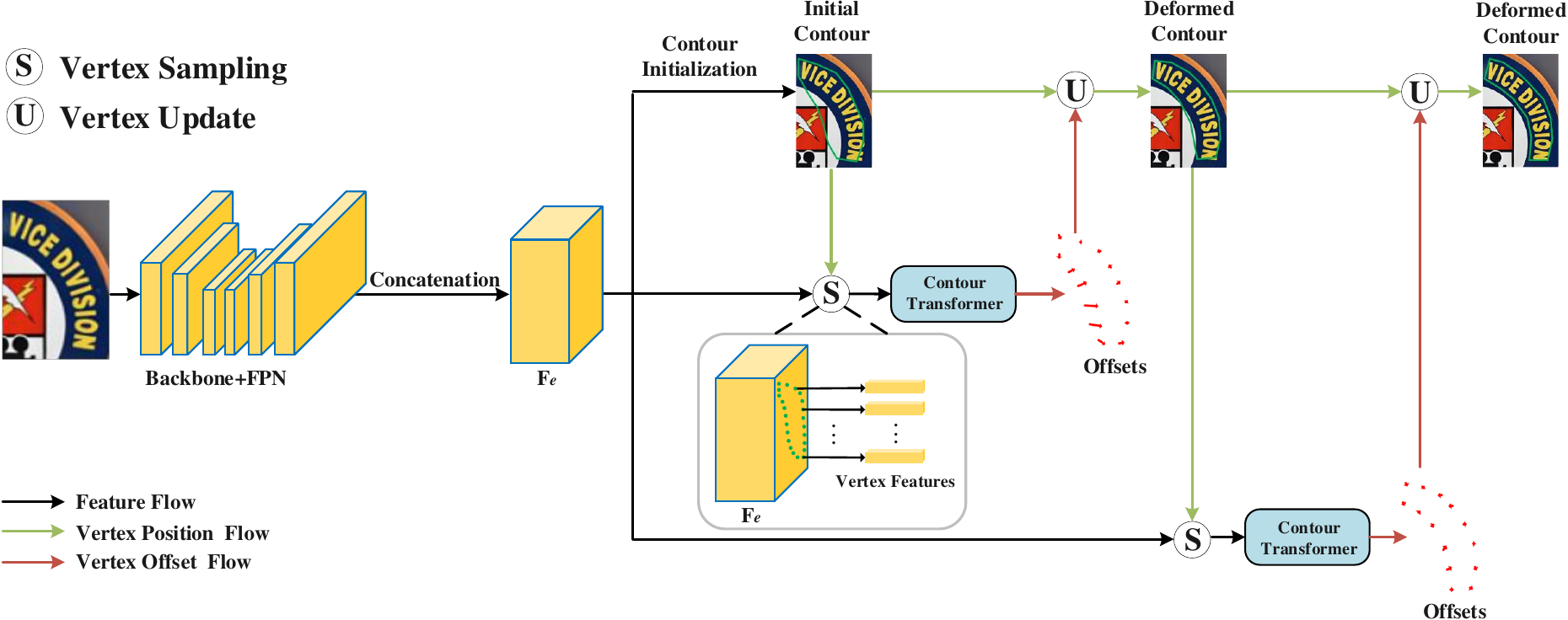}
\caption{
The architecture of our CN-Net, which is composed of three modules: feature extraction by the ResNet50\cite{he2016deep} and FPN \cite{lin2017feature}, contour initialization module for initial contour generation, and contour refinement modules for progressive contour deformation. Two stages of contour refinement are illustrated. \highlight{In each stage, the feature of each vertex token is sampled from $\mathbf{F}_e$ according to its spatial position specified by the contour.}
% The architecture of our CT-Net. Given an input image, we first extract multi-scale features by FPN \cite{lin2017feature}, where the backbone is ResNet50 \cite{he2016deep}. Then we fuse these features to obtain a representative feature ${F}_e$. Next, ${F}_e$ is fed into the contour initialization module to generate initial contours. Finally, we employ contour refinement modules for progressive contour deformation.
}
\label{fig-overview}
\end{figure*}

\highlight{\textit{Second, error accumulation exists between different stages, e.g., imprecise detection results in the previous stage may hinder the estimation in the current stage.}} Two reasons mainly cause this. On the one hand, the contour refinement module cannot effectively evaluate the confidence of the contours to solve the false positive problem. Although PCR \cite{dai2021progressive} proposes a reliable contour localization mechanism (RCLM) to evaluate the confidence of the contours, its negative sample mining technique does not generate text-like negative samples. On the other hand, the contour refinement module learns only the fixed contour deformation path between the ground-truth initial contour and the ground-truth final contour \highlight{during training}, which makes it difficult to obtain its refinement direction when the predicted contour deviates from the ground-truth \highlight{during inference}. \highlight{Besides}, the tasks of initial contour prediction and progressive contour refinement are independent of each other, which may not maximize the final performance of text detection.

\highlight{\textit{Third, local information aggregation mechanism ignores the access of global information.}} \highlight{Such local information aggregation often uses} circular convolution \cite{peng2020deep} or graph convolution \cite{kipf2016semi} to propagate the features of adjacent contour vertexes. However, when the prediction of individual contour vertex has a large deviation, it is difficult to accurately perceive the refinement direction only through the information of adjacent contour vertexes.
% is extensively applied in contour adjustment, which refines the contours by  through . However, this type %repetition
% of local information aggregation to access global information is inefficient.
% It cannot correct large prediction errors because it is difficult to accurately perceive the optimization direction only through the information of the local contour vertexes around them .
Moreover, as the number of contour vertexes increases, the global information becomes more difficult to access, leading to a decrease in detection performance, while a small number of contour vertexes have limited capability to completely fit arbitrary-shaped text boundaries.

In this paper, we propose a novel contour transformer based scene text detection framework called \textbf{CT-Net}, which solves the above problems through three components: 1) a contour initialization module; 2) an adaptive training strategy with a re-score mechanism; and 3) transformer-based global contour refinement modules.
In particular, the contour initialization module produces bounding polygons as the initial contours based on features of positive points, which can roughly represent the text contours, as shown in Fig.~\ref{fig-basic}(a). It introduces negligible computation overhead compared with typical bounding box detection.
%and controls the granularity of the regression by adaptively adjusting the number of vertexes of the boundary polygons, which introduces negligible computational overhead compared to the object detection task.
Then, the coarse initial contour is deformed by the contour refinement module, as shown in Fig.~\ref{fig-basic}(b) and (c).
%It fully excavates the global contextual information of the contours through a self-attentive mechanism, and then performs the global contour deformation by a multilayer perceptron network (MLP).

\highlight{Inspired by the vision transformers (ViTs)~\cite{dosovitskiy2020image,xu2023spatiotemporal,xu2023pyramid},} the contour refinement module aggregates the global information by a multi-layer perceptron (MLP) to achieve the prediction of larger offsets, and excavates contextual information by a self-attention mechanism to learn the overall layout of the contours. The deformed vertexes are able to enclose closed curves in order, which significantly reduces the difficulty of contour deformation. Meanwhile, to allow the contour refinement module to learn more potential deformation paths, we propose an adaptive training strategy to enable the contour refinement module can effectively refine the predicted initial contours during training. Considering errors may be accumulated by initial contours of false detection, we propose a re-score mechanism to further reduce the error accumulation by evaluating the confidence of the predicted contour based on the contour vertex features. The joint optimization of initial contour prediction and progressive contour deformation is achieved by the adaptive training strategy and the re-score mechanism, in which the process of contour deformation is forced to be robust to contour initialization.
% and forces the latter more robust to the former.

%Moreover, most of the existing text detectors are confused by the text-like background, To solve this problem, we propose a Re-Score mechanism that reclassifies the text-like background.

The main contributions of this paper are summarized as follows:
\begin{itemize}
    \item We propose a novel multi-stage contour based framework for arbitrary-shaped scene text detection, which can accurately and efficiently localize text contours without any post-processing.
    \item We propose \highlight{a novel initial contour representation method with polygons to decrease the training complexity, as well as novel} contour refinement modules to adaptively refine text contours in an iterative manner.
    \item We design an adaptive training strategy with a re-score mechanism to \highlight{jointly optimize initial contour prediction and progressive contour refinement, which is beneficial for reducing} error accumulation across stages.
    \item Extensive experiments show that our CT-Net achieves state-of-the-art performance. Specifically, CT-Net obtains $86.1$ F-measure at $11.2$ frames per second (FPS) and $87.8$ F-measure at $10.1$ FPS on CTW1500~\cite{liu2019curved} and Total-Text~\cite{ch2017total}, respectively.

\end{itemize}

\section{Related Work}
With the booming development of deep learning techniques, numerous inspiring ideas and effective methods for scene text detection tasks have been investigated. In this section, we will review the related methods on this topic.

\subsection {Segmentation Based Methods}
Inspired by semantic segmentation methods \cite{long2015fully,ronneberger2015u}, some works regard the text detection as a segmentation problem, and then separate different text instances through heuristic post-processing. TextSnake \cite{long2018textsnake} regarded text instances as a series of overlapping disks, and predicted some geometric attributes of the disks to reconstruct the text instances. PSENet \cite{wang2019shape} predicted different scale kernels of text instances, then adopted a progressive scale expansion strategy to gradually expand the predefined text kernels. TextField \cite{xu2019textfield} learnt a deep direction field that encodes both text mask and direction information of each point, then linked neighbor pixels to generate candidate text instances. SAE \cite{tian2019learning} treated each text instance as a cluster and adopted a two-step clustering strategy to distinguish adjacent text instances from the predicted center and full segmentation maps. Moreover, DBNet \cite{liao2020Real} proposed a differentiable binarization module, which gives a higher threshold to the text boundaries to distinguish adjacent text instances. %Unsatisfactory,
However, these segmentation based methods are sensitive to the text-like background noises and lack global perception of the texts, resulting in many noises and defects in the predicted contours.

\subsection {Regression Based Methods}
Regression based methods rely on the object detection framework with box regression, which are easier to train than segmentation based methods.
RRPN \cite{ma2018arbitrary} used a modified pipeline of Faster R-CNN \cite{ren2015faster}, which adopted rotated bounding boxes to represent multi-oriented texts. TextBoxes \cite{liao2017textboxes} modified the shapes of convolutional kernels and increased the scale of the default boxes of SSD \cite{liu2016ssd} to fit the aspect ratio of texts. TextBoxes++ \cite{liao2018textboxes++} extended TextBoxes by using quadrilateral regression to detect multi-oriented texts. EAST \cite{zhou2017east} and MOST \cite{he2021most} adopted a single-stage anchor-free framework to directly regress the offsets between points within text instances and the corresponding boundaries or vertexes.
%The two methods achieve a better trade-off between efficiency and accuracy due to the elimination of anchor effects. However, the above regression based text representations are horizontal or oriented rectangle, which have limited capacity to model irregular texts.
To effectively detect curved texts,
LOMO \cite{2019Look} introduced a shape representation module that uses text regions, center lines and boundary offsets to represent text instances, and proposed an iterative refinement module to detect long texts. Tang \etal\cite{tang2022few} first obtained representative text region features by feature sampling and then used the feature aggregation capability of the self-attention mechanism to locate texts accurately. However, these methods either suffer from error accumulation in multi-stage or cannot sufficiently adapt to arbitrary-shaped text detection.

\subsection {Connected Component Based Methods}
Connected component based methods typically first detect individual text parts, then generate final detection through a group post-processing procedure. CTPN \cite{tian2016detecting} first %used a modified framework of Faster R-CNN \cite{ren2015faster} to
extracted a series of text components with a fixed width, then connected them into horizontal text lines. %generating horizontal text lines.
CRAFT \cite{baek2019character} adopted two-dimensional Gaussian segmentation labels to detect character-level text regions, then generated final detection based on the affinity between characters.
DRRG \cite{2020Deep} first predicted the geometry attributes of text components via a text proposal network, then used a relational reasoning graph network to infer the linkages between the text components. Although these methods allow flexible representation of arbitrary-shaped texts, the complex post-processing of text component clustering seriously affects the detection efficiency.

\subsection {Contour Based Methods}
Contour based methods can directly obtain the text contours without complicated post-processing.
TextRay \cite{wang2020textray} formulated the text contours in the polar system, and adopted the Chebyshev polynomials to approximate the text contours. ABCNet \cite{liu2020abcnet} proposed a single-shot anchor-free framework to predict a series of Bezier points, then adopted Bernstein polynomial to convert these points into Bezier curves that approximate the text contours.
FCENet \cite{zhu2021fourier} modeled text instances in the Fourier domain, and adopted discrete Fourier transform to fit the text contours.
To obtain more accurate text contours,
TextBPN \cite{zhang2021adaptive} first obtained boundary proposals based on a bottom-up approach, then gradually deformed the boundary proposals into text contours through multiple adaptive boundary deformation modules. PCR \cite{dai2021progressive} proposed a progressive contour regression
approaches to detect arbitrary-shaped scene texts from a top-down perspective. However, these single-stage methods \cite{wang2020textray,liu2020abcnet,zhu2021fourier} have limited ability to model extremely long texts or highly curved texts, while these multi-stage methods \cite{zhang2021adaptive,dai2021progressive} suffer from the error accumulation.

\begin{figure}
\centering
\includegraphics[width=\linewidth]{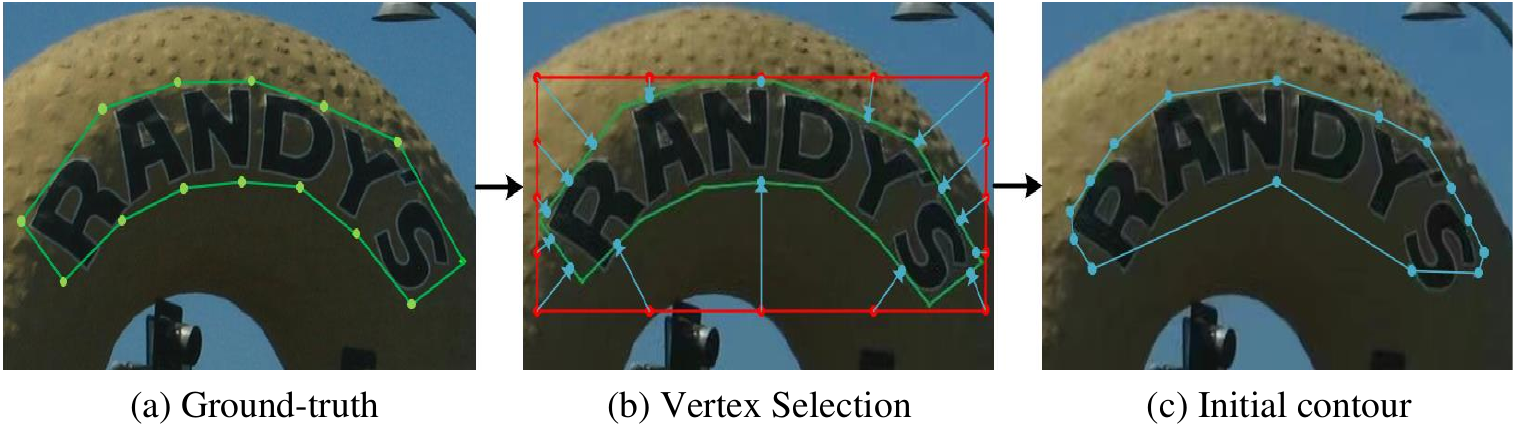}
\caption{The generation of initial contours. (a) The ground-truth polygon boundary. (b) The sampling on boundary. (c) The \highlight{ground-truth} initial contour composed of sampling vertexes.
}
\label{fig-initialcontour}
\end{figure}

\section{Arbitrary-Shaped Text Detection via Contour Transformer}

\subsection{Overview}
Our CT-Net is a multi-stage contour based detection framework that can fit arbitrary-shaped texts. %in any scenario.
As illustrated in Fig.~\ref{fig-overview}, this framework mainly contains three parts: feature extraction module, contour initialization module, and multiple contour refinement modules. In the feature extraction module, we utilize ResNet50 \cite{he2016deep} as the backbone and adopt FPN \cite{lin2017feature} to generate multi-scale features, and then fuse these \highlight{multi-scale features by concatenation} to obtain a contextual feature ${F}_e$. In the contour initialization module, we adopt a single-shot anchor-free framework to predict the initial contour.
As shown in Fig.~\ref{fig-overview}, we adopt multiple contour refinement modules to iteratively refine the predicted contours in each contour refinement module, we first utilize multiple transformer encoders to aggregate the global information of the text contours and then predict the offset and confidence of the text contours by an MLP, respectively.
% in a same way as \cite{wang2019shape}

\subsection {Contour Initialization}
\label{ssec:contour_init}

%A good contour initialization module should be able to locate the text directly in a concise and efficient manner with low training complexity. In this paper, we use a fixed number of polygon vertexes to locate the approximate position of the text. and obtain these vertexes by an anchor-free framework.

Typical single-stage contour based methods like ABCNet \cite{liu2020abcnet} and FCENet \cite{zhu2021fourier} use the Bezier point and Fourier coefficient to represent text contours. However, their parameters fluctuate greatly, which makes training difficult.
% their parameters vary drastically from one another, which makes training difficult, especially when the number of parameters is large.
Instead, we propose a contour representation method based on bounding polygon vertexes, which have a good trade-off between training complexity and contour representation quality.

Specifically, as shown in Fig.~\ref{fig-initialcontour}(a), the ground-truth polygon boundary of a scene text contains $N$ points $\{(\bar{x}_{1}, \bar{y}_{1}),(\bar{x}_{2}, \bar{y}_{2}),\cdots,(\bar{x}_{N}, \bar{y}_{N})\}$ with vertexes as well as points in each edge. Firstly, we
%focus on the ground-truth bounding box instead of ground-truth polygon. We
uniformly sample $N_a$ vertexes $\{(\hat{x}_{1}, \hat{y}_{1}),(\hat{x}_{2}, \hat{y}_{2}),\cdots,(\hat{x}_{N_a}, \hat{y}_{N_a})\}$ from the four edges of the ground-truth bounding box in clockwise order, in which the bounding box can be determined from the ground-truth polygon. Then, we sample $(x_{i}, y_{i})$ by choosing the point in the ground-truth polygon with the nearest L1 distance to $(\hat{x}_{i}, \hat{y}_{i})$:
\begin{equation}
\label{minxy}
\left(x_{i}, y_{i}\right) = \arg \underset{(\bar{x}_{j}, \bar{y}_{j})}{\min } \sum_{j=1}^{N} (\left|\bar{x}_{j} -\hat{x}_{i}\right| + \left|\bar{y}_{j} -\hat{y}_{i}\right|),
\end{equation}
where $i=1,2, \cdots, N_a$, and $N_a$ is usually much smaller than $N$.

% from $(\bar{x}_{j}, \bar{y}_{j}), j=1,2, \cdots, N_a$, according to the from each point in $(\bar{x}_{j},\bar{y}_{j})$ to $(\hat{x}_{k}, \hat{y}_{k})$:

% the generation process of the  vertexes  consists of three steps, $k=1,2, \cdots, . Firstly, , we sample $(\hat{x}_{i}, \hat{y}_{i})$ on the ground-truth bounding box, $i=1,2, \cdots, N_a$, in which vertexes on the

% . Secondly, as shown in Fig.~\ref{fig-ct}(a), we generate the text contour vertexes $(\bar{x}_{j}, \bar{y}_{j})$ in clockwise order, pixel by pixel, $j=1,2, \cdots, N_a$, according to the ground-truth text contour. Thirdly,

Since $(\hat{x}_{i}, \hat{y}_{i})$ can be directly determined from the bounding box, we propose to regress the position of the bounding box as well as the horizontal-vertical offset between each pair of $(x_{i}, y_{i})$ and $(\hat{x}_{i}, \hat{y}_{i})$. In this way, our method is easier to train, as the bounding box is less complex than the polygon and the fluctuations of the minimum distances between sampled points are small.

In our contour initialization module, we first use three convolutions to enhance features, then use one convolution to predict the probability of each position corresponding to the positive sample, the regression value of each position corresponding to the text bounding box, and the vertex offsets, respectively. To achieve end-to-end prediction without non-maximal suppression (NMS), we select only one positive sample for each text instance during training. Inspired by \cite{sun2021makes}, our positive sample allocation strategy is as follows: in each training iteration, we select only one positive sample that can minimize the training loss for each text instance, and the training loss is defined as the sum of classification loss and regression loss.

During inference, there are three steps for the prediction of initial contour. Firstly, we obtain the text box based on the classification threshold $\tau_a$ and the regression value of the box. Secondly, we sample the predicted text box according to the sampling strategy during training. Thirdly, the initial contour vertexes are obtained based on the position of the sampled points and their corresponding offsets.

\begin{figure}
\centering
\includegraphics[width=\linewidth]{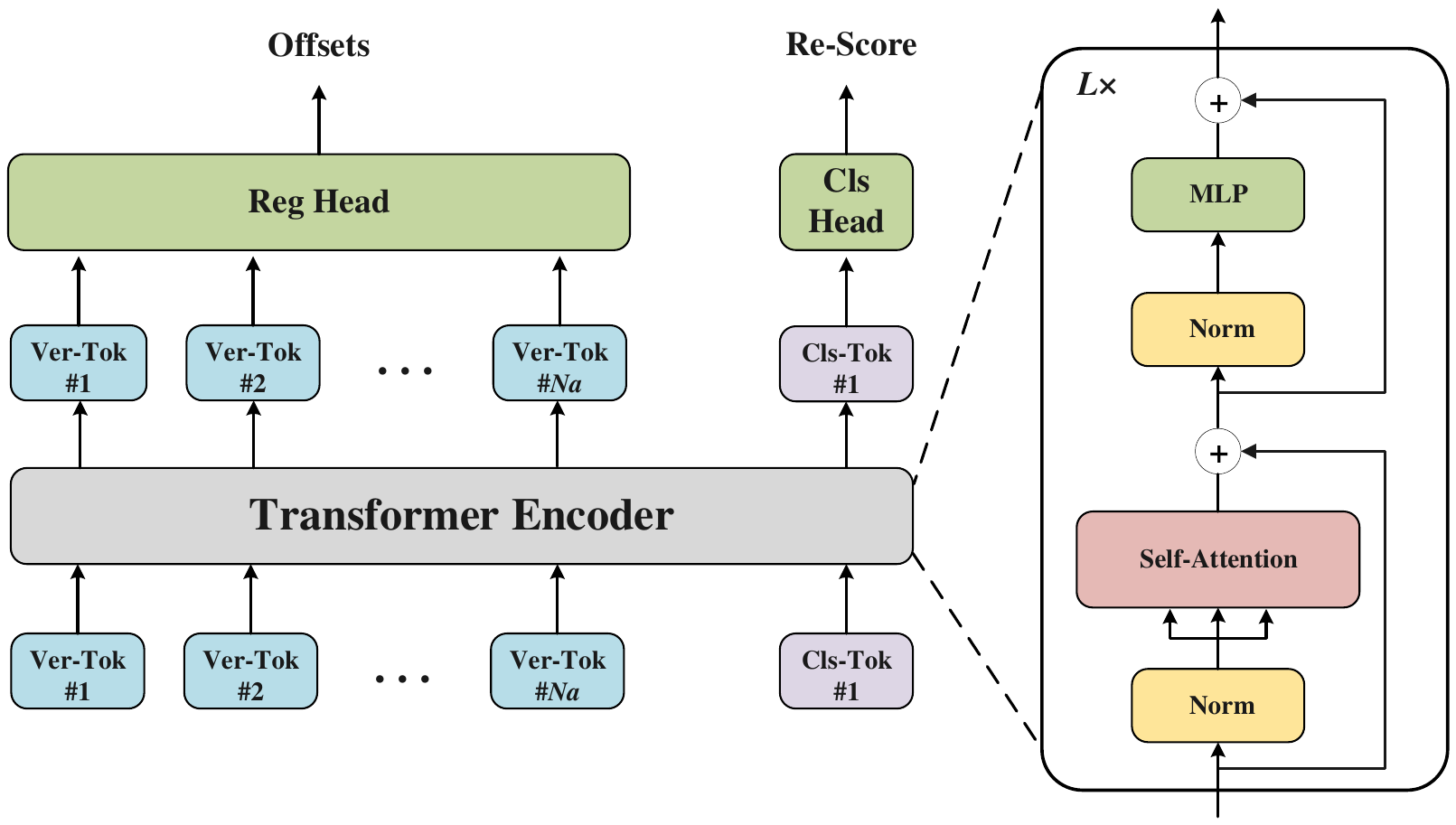}
\caption{The detailed structure of a contour refinement module. ``Ver-Tok'' refers to [PATCH] token, which is the feature of the contour vertex corresponding to ${F}_e$. ``Cls-ToK'' refers to [CLS] token, which is a learnable embedding for contour confidence. %$L$ refers to the number of transformer encoder.
}
\label{fig-ct}
\end{figure}

\subsection {Contour Transformer}

In this module, we first initialize the contour with $N_a$ vertexes, and then employ multiple contour refinement modules to progressively refine the initial contours, as shown in Fig.~\ref{fig-overview}.

Fig.~\ref{fig-ct} elaborates the architecture of a contour refinement module. The main difference between ViT \cite{dosovitskiy2020image} and our contour transformer is the prediction head. Specifically, the input of our contour transformer is
$\mathbf{z}_{0}=\left[ \mathbf{x}_{ver}^{1}; \mathbf{x}_{ver}^{2}; \cdots ; \mathbf{x}_{ver}^{N_a}; \mathbf{x}_{\text {cls }} \right]$.
Here, ${\mathbf{x}}_{ver}^{i}\in{\mathbb{R}^{1 \times C}}$ denotes the feature of the $i$-th contour vertex token, \highlight{which is sampled from $F_e$ according to its spatial position, as shown in Fig.~\ref{fig-overview}.} $C$ is the number of input channels.
%Here, ${\mathbf{x}}_{ver}^{i}\in{\mathbb{R}^{1 \times C}}$ denotes the feature of the $i$-th contour vertex corresponding to ${F}_e$, where $C$ is the number of input channels.
Considering that the contours may be evolved from some false positives, we employ a learnable classification token ${\mathbf{x}}_{\text {cls }}\in{\mathbb{R}^{1 \times C}}$ to learn the confidence of the detected contours (see more details in Sec.~\ref{ssec:re-socre}).

%the input is sent to multiple transformer encoder to efficiently learn the global context information of contour vertexes.
The body of our contour transformer is composed of $L$ transformer encoder layers for global information interaction between contour vertexes. Each transformer encoder layer consists of a multi-head self-attention (MSA) block and an MLP block. LayerNorm (LN) \cite{ba2016layer} is applied before each block and residual connection is applied after each block. The MLP block is made of two layers with a non-linearity activation function GELU \cite{hendrycks2016gaussian}. The $\ell$-th transformer encoder layer can be formulated as
\begin{equation}
\begin{aligned}
&\mathbf{z}_{\ell}^{\prime}=\operatorname{MSA}\left(\operatorname{LN}\left(\mathbf{z}_{\ell-1}\right)\right)+\mathbf{z}_{\ell-1}, \\
&\mathbf{z}_{\ell}=\operatorname{MLP}\left(\operatorname{LN}\left(\mathbf{z}_{\ell}^{\prime}\right)\right)+\mathbf{z}_{\ell}^{\prime} .
\end{aligned}
\end{equation}

The decoder of our contour transformer consists of a regression (Reg) head and a classification (Cls) head, both of which are implemented by an MLP with separate parameters. Both heads have two hidden layers and the former two fully connected layers are equipped with the ReLU-BN-Dropout operation, where the dropout probability is $0.1$ and the channels of hidden layers are $C$. The only difference between the two heads is the channels of the output layer ($N_a \times 2$ for the regression head and $1$ for the classification head). For regression head, the output is the offset of the contour vertexes (a vector of length $N_a \times 2$). The offsets and the positions of input features are summed to obtain the deformed contour.

\subsection {Adaptive Training Strategy}

The contour refinement module of current multi-stage contour based approaches \cite{wu2021progressive,zhang2021adaptive} learns only the deformation from the ground-truth initial contour to the ground-truth final contour \highlight{during training},
% ground-truth vertexes,
which may lead to some errors when the predictions of the initial contour are not accurate \highlight{during inference}. %as the ground-truth initial contour.
To alleviate such issues, we propose an %simple yet effective method, named
adaptive training strategy for learning more potential deformation paths. %by putting contour predictions into the training of deformation process.
% considering intermediate predictions in the deformation process.

Specifically, we make our contour transformer learn not only the offset between the ground-truth initial contour and the ground-truth final contour, but also the offset between the predicted initial contour and the ground-truth final contour. Formally, the predicted text bounding boxes $\mathbf{B}^f$ and the corresponding predicted contours $\mathbf{C}^f$ in the contour initialization module are obtained in the same way as the inference stage (see Sec.~\ref{ssec:contour_init}).
%The core of adaptive training strategy is to assign labels to $\mathbf{C}^f$.
% The corresponding
% ground-truth of the contour refinement module is then assigned based on
First, the ground-truth index of \highlight{the $i$-th predicted contour} $C^{f}_{i}$ is obtained by:
\begin{equation}
i n d_{i}= \arg\underset{k \in[1 \cdots, N_{b}]}{ \max}\operatorname{IOU}\left(B_{i}^{f}, B_{k}^{gt}\right),
\end{equation}
where $\mathbf{B}^{gt}$ denotes totally $N_{b}$ ground-truth bounding boxes, and IOU denotes intersection over union. %, and $N_{b}$ is the number of the ground-truth bounding boxes.
% After assigning the annotation to the $C^f$, we simply concatenate the new target to the original ground-truth set for further contour refinement training.
% In this way, we can obtain the ground-truth label of $C^f$, which
Then, $\mathbf{C}^f$ is used for training contour deformation. Note that there may be some false positives in $\mathbf{C}^f$, \highlight{as shown in the right side of Fig.~\ref{fig-rescore}}.
We \highlight{discard} these false positives for the training of the regression head in the contour transformer, \highlight{since they do not have regression targets}. $C^{f}_{i}$ is simply regarded as a false positive when: %$iou_{i}$ is less than $0.5$:
\begin{equation}
\label{iou}
i o u_{i}= \underset{k \in[1 \cdots, N_{b}]}{ \max}\operatorname{IOU}\left(B_{i}^{f}, B_{k}^{gt}\right)<0.5.
\end{equation}
Besides the contour initialization module, this adaptive training strategy is also used in each contour refinement module.

\subsection {Re-Score Mechanism}
\label{ssec:re-socre}

To effectively evaluate the confidences of the detected contours, we exploit a re-score mechanism in parallel with the contour deformation. Specifically, we first add a learnable $\mathbf{x}_{\text {cls }}$ token to the input of the contour refinement module to learn the category information of contour vertexes, then extract the category information of the $\mathbf{x}_{\text {cls }}$ token through a classification head to generate contour scores for text/non-text.

\begin{figure}
\centering
\includegraphics[width=\linewidth]{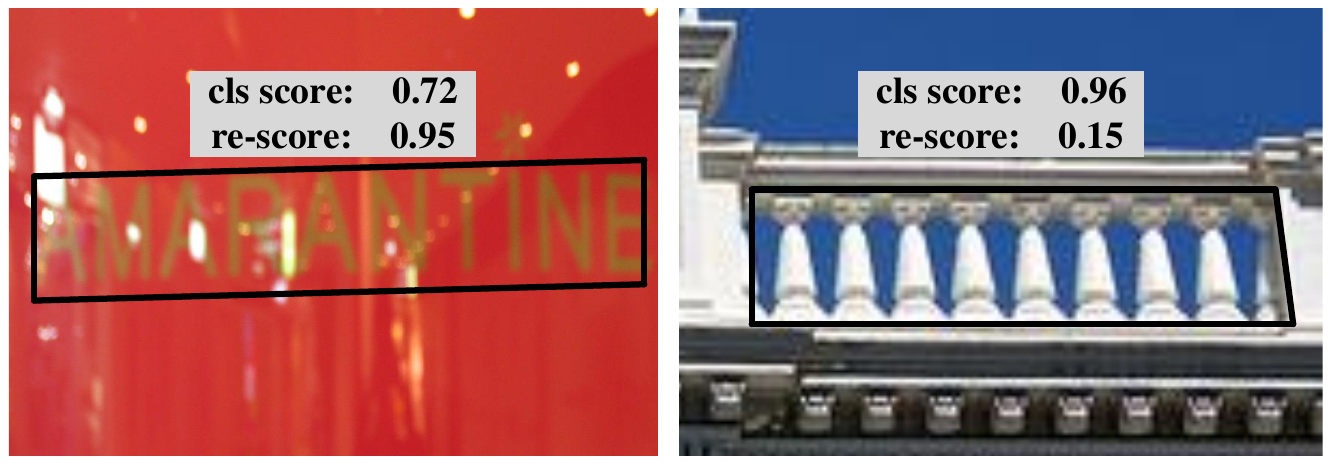}
\caption{The visualization of detection results with the classification score and the re-score.
}
\label{fig-rescore}
\end{figure}

To learn a robust contour scoring network, it requires samples with different labels to train this network for distinguishing the contours of scene texts from those of backgrounds. In particular, the input of the contour refinement module comes from two parts, 1) the features corresponding to the ground-truth contour and 2) the features obtained by the adaptive training strategy. For the classification head of the contour refinement module, we define the classification labels of the former as $1$, while the latter has many \highlight{false positives}, in which we adopt $iou_i$ in Eq.~\eqref{iou} to represent the confidence of these samples as text contours and regard them as classification labels. With the continuous optimization of the model in training, \highlight{false positives} will be significantly reduced. To improve the model's ability of distinguishing false positives, we adopt quality focal loss (QFL) \cite{li2020generalized} to increase the loss weights of the false positives. As shown in Fig.~\ref{fig-rescore}, re-score mechanism can significantly reduce the classification score of the false positives.
%\highlight{the $C^{f}$ obtained by adaptive training strategy will also be gradually accurate. Therefore, false positives in $C^{f}$ will be gradually reduced.}

Moreover, since our re-score mechanism can accurately evaluate the contour confidence, we propose an early-stop mechanism that allows our model to stop itself when it has enough confidence score. Specifically, we only send contours with a confidence score below the classification threshold $\tau_b$ to the next contour refinement module for contour deformation.

\subsection{Full Loss Function}
In our CT-Net framework, the full loss function is formulated as
\begin{equation}
\label{loss1}
\mathcal{L}=L_{init}+ L_{transform},
\end{equation}
where $L_{init}$ and $L_{transform}$ are the losses for the initial contour module and the contour refinement module, respectively.

In Eq.~\eqref{loss1}, $L_{init}$ is computed as
\begin{equation}
\label{loss2}
L_{init}=L_{cls}+\lambda_{1} L_{box} + \lambda_{2} L_{off1},
\end{equation}
where $L_{cls}$, $L_{box}$, and $L_{off1}$ denote classification loss, bounding box regression loss, and offset regression loss, respectively. We choose cross-entropy loss to optimize $L_{cls}$, choose smooth-L1 loss \cite{ren2015faster} to optimize $L_{off1}$, and use GIoU loss \cite{rezatofighi2019generalized} for $L_{box}$ following ABCNet \cite{liu2020abcnet}.
$L_{transform}$ is computed as
\begin{equation}
\label{loss3}
L_{transform}=\lambda_{3} L_{off2}+\lambda_{4} L_{rescore},
\end{equation}
where $L_{off2}$ and $L_{rescore}$ denote contour deformation loss and re-score loss, respectively. We choose QFL \cite{li2020generalized} to optimize $L_{rescore}$, and define the contour deformation loss as
\begin{equation}
L_{off2}=\min _{u \in[1 \cdots, N_a]} \sum_{i=1}^{N_a} \operatorname{smooth-L1} \left(p_{i}, p_{(u+i)\%  N_a}^{\prime}\right),
\end{equation}
where $\mathbf{p}$ and $\mathbf{p}^{\prime}$ denote the prediction and ground-truth contour vertex sets, respectively, and ``$\%$'' denotes the remainder operation.

\section{Experiments}

\subsection{Datasets and Settings}
\subsubsection{Datasets}
We evaluate our CT-Net on four scene text detection benchmarks: CTW1500~\cite{liu2019curved}, Total-Text~\cite{ch2017total}, ICADAR2015~\cite{karatzas2015icdar}, and MSRA-TD500~\cite{yao2014unified}.
\begin{itemize}
    \item \textbf{CTW1500} is a challenging dataset that contains many long curved texts. It consists of $1,000$ training images and $500$ test images, in which each text is annotated at text-line level with $14$ vertexes.

    \item \textbf{Total-Text} includes $1,255$ images for training and $300$ images for testing. Texts in this dataset have various shapes, including horizontal, oriented, and curved shapes. All text instances are annotated with adaptive number of contour vertexes.

    \item \textbf{ICDAR2015} is presented in the ICDAR 2015 Robust Reading Competition with $1,000$ images for training and $500$ images for testing. Texts in this dataset are all annotated with four vertexes of quadrangle.

    \item \textbf{MSRA-TD500} is a scene text dataset for detecting multi-orientated long texts. It consists of $300$ images for training and $200$ images for testing,  including English and Chinese texts. Besides, the annotations of these datasets are all in line-level.
\end{itemize}

\subsubsection{Implementation Details}
Our CT-Net is implemented via PyTorch~\cite{paszke2019pytorch}. The backbone network ResNet50 \cite{he2016deep} is pre-trained
on ImageNet \cite{krizhevsky2012imagenet}.
The training of CT-Net can be divided into two steps. First, we pre-train our network on a large synthetic dataset SynthText~\cite{gupta2016synthetic} with totally $800,000$
images by four epochs. Then, we fine-tune our model on specific datasets with a mini-batch size of $4$. The model is trained up for %$120,000$ iterations
$480$ epochs with the initial learning rate of $1\times10^{-4}$, in which the learning rate is decreased to $1\times10^{-5}$ at the $360$-th epoch. %$90,000$-th iteration.
Following other methods~\cite{liu2020abcnet,liao2020Real,wang2019efficient}, the text regions labeled as ``DO NOT CARE'' are ignored during training. Moreover, we set $N_a$ to $32$, $L$ to $4$, $\tau_a$ to $0.45$, and $\tau_b$ to $0.5$. In Eq.~\eqref{loss2}, $\lambda_1=1$, and $\lambda_2=1$. In Eq.~\eqref{loss3}, $\lambda_3=0.1$, and $\lambda_4=2$.

For data augmentation, we randomly perform horizontal flip, rotation, scale, crop, color jitter, and contrast jitter for training images. In random scale, the short side of input images is uniquely chosen from $640$ to $832$ with an interval of $32$, and the long side is less than $1,600$ for CTW1500, Total-Text, and MSRA-TD500. For ICDAR2015, the short side of input images is uniquely chosen from $980$ to $1432$ with an interval of $32$, and the long side is less than $2,800$. In random crop, we only crop the non-text regions, in which the crop area is larger than half of the original image area. All experiments are carried out on an NVIDIA Tesla V100 GPU. \highlight{The average running speeds (FPS) of different methods during inference are all computed by feeding a single image into the model at a time.} We employ three popular evaluation metrics recall (R), precision (P), and F-measure (F), in which $\%$ in their results is omitted for simplicity. %to evaluate our method.

\begin{table}[]
\setlength\tabcolsep{8pt}
\centering
\caption{Results for different variants of our CT-Net on CTW1500 \cite{liu2019curved} without pre-training, in which CR and AT denote the contour refinement module and the adaptive training strategy, respectively.}
\label{tabel-ablation-1}
\begin{tabular}{c|ccc|ccc}
\hline
\multirow{2}{*}{Method} &\multicolumn{3}{c|}{Components}         & \multicolumn{1}{c}{\multirow{2}{*}{R}} & \multicolumn{1}{c}{\multirow{2}{*}{P}} & \multirow{2}{*}{F} \\ \cline{2-4}
  &CR & AT & \multicolumn{1}{c|}{re-score} & \multicolumn{1}{c}{}                       & \multicolumn{1}{c}{}                       &                        \\ \hline
BNet &-  & -        & -                        & $81.9$                                           & $82.8$                                           & $82.3$                      \\
BCNet &\checkmark  & -        & -                        & $83.1$                                           & $84.9$                                           & $84.0$                      \\
BCANet &\checkmark  & \checkmark        & -                        & $\mathbf{83.7}$                                           & $85.2$                                           & $84.5$                      \\
BCRNet &\checkmark  & -        & \checkmark                        & $81.9$                                           & $87.7$                                           & $84.7$                      \\
% \tabincell{c}{CT-Net\\w/o pre-train}
\textbf{CT-Net}&\checkmark  & \checkmark        & \checkmark                        &  $82.7$                                       & $\mathbf{87.9}$                                          & $\mathbf{85.2}$                   \\ \hline
\end{tabular}
\end{table}

% \begin{table}[]
% \renewcommand\arraystretch{1.2}
%  \centering
%  \caption{Results for different variants of contour refinement module on CTW1500~\cite{liu2019curved} and Total-Text~\cite{ch2017total}.}
%  \label{tabel2}
% \begin{tabular}{c|c|ccc}
% \hline
% Dataset    & Method         & R        & P    & F    \\ \hline
% \multirow{3}{*}{CTW1500}    & CCN \cite{peng2020deep} & $82.4$          & $86.9$          & $84.6$          \\
%           & GCN \cite{kipf2016semi} & $82.5$          & $87.0$          & $84.7$          \\
%           & \textbf{Contour transformer}        & $\mathbf{82.7}$ & $\mathbf{87.9}$ & $\mathbf{85.2}$ \\ \hline
% \multirow{3}{*}{Total-Text} & CCN \cite{peng2020deep} & $82.6$          & $88.8$          & $85.6$          \\
%           & GCN \cite{kipf2016semi} & $83.1$          & $88.7$          & $85.8$          \\
%           & \textbf{Contour transformer}        & $\mathbf{83.6}$ & $\mathbf{89.2}$ & $\mathbf{86.3}$ \\ \hline
% \end{tabular}
% \end{table}
\begin{table}[]
\renewcommand\arraystretch{1.4}
\setlength\tabcolsep{5.5pt}
 \centering
 \caption{\highlight{Complexities and} Results for different variants of contour refinement module on CTW1500~\cite{liu2019curved} and Total-Text~\cite{ch2017total}.}
 \label{tabel2}
\begin{tabular}{c|c|ccc|ccc}
\hline
\multicolumn{2}{c|}{Method} &\highlight{\tabincell{c}{Params\\($\times 10^6$)}} &\highlight{\tabincell{c}{FLOPs\\($\times 10^9$)}}  &\highlight{FPS}    & R        & P & F    \\ \hline
\multirow{3}{*}{\rotatebox[origin=c]{90}{CTW1500}}    & CCN \cite{peng2020deep} &$2.85$ & $1.54$   & $10.6$         & $82.4$          & $86.9$ & $84.6$          \\
           & GCN \cite{kipf2016semi} & $2.71$ & $1.38$ & $10.9$          & $82.5$          & $87.0$ & $84.7$          \\
           & \textbf{CT}        & $\mathbf{2.55}$ & $\mathbf{1.28}$  &$\mathbf{11.3}$ & $\mathbf{82.7}$ & $\mathbf{87.9}$ & $\mathbf{85.2}$ \\ \hline
\multirow{3}{*}{\rotatebox[origin=c]{90}{Total-Text}} & CCN \cite{peng2020deep} & $2.85$ & $1.54$  &$9.3$ & $82.6$          & $88.8$ & $85.6$          \\
           & GCN \cite{kipf2016semi} &$2.71$& $1.38$ &$9.6$  & $83.1$          & $88.7$   & $85.8$          \\
           & \textbf{CT}        &$\mathbf{2.55}$ &$\mathbf{1.28}$ &$\mathbf{9.8}$ & $\mathbf{83.6}$ & $\mathbf{89.2}$ & $\mathbf{86.3}$ \\ \hline
\end{tabular}
\end{table}

\subsection{Ablation Study}
In this section, we conduct ablation study on the CTW1500 dataset. %, in which these variants of our CT-Net are not pre-trained on SynthText~\cite{gupta2016synthetic} for convenience. %are trained directly on the training set of CTW1500.
% To verify the advantage of our CT-Net, w
We introduce a BNet as the baseline whose output comes only from the contour initialization module. BNet is not pre-trained on SynthText~\cite{gupta2016synthetic} for simplicity.

\subsubsection{Contour Transformer}

As shown in Table~\ref{tabel-ablation-1}, when BCNet integrates
the contour refinement module over BNet,
it improves $1.7$ in F-measure. The improvement %could be ascribed
is due to more accurate localization by multiple perception on scene texts with complex geometric layouts.
To further verify the effectiveness of our contour transformer, we compare it with the previous deformation methods, circular convolution network
(CCN) \cite{peng2020deep} and graph convolution network (GCN) \cite{kipf2016semi}. For a fair comparison, we only replace the body of our contour transformer (CT) with CCN or GCN in the contour refinement module.

\begin{table}[]
\setlength\tabcolsep{22pt}
\centering
\caption{Performance of CT-Net with different \highlight{stages} of contour transformers on CTW1500~\cite{liu2019curved}.}
\label{tabel-ablation-2}
\begin{tabular}{c|ccc}
\hline
\highlight{Stage}   &  1 &  \textbf{2}        &  3        \\ \hline
F  & 84.8   & \textbf{85.2} & \textbf{85.2} \\
FPS   & \textbf{13.3}   & 12.2          & 11.2  \\
\hline
\end{tabular}
\end{table}

\begin{figure}
\centering
\includegraphics[width=\linewidth]{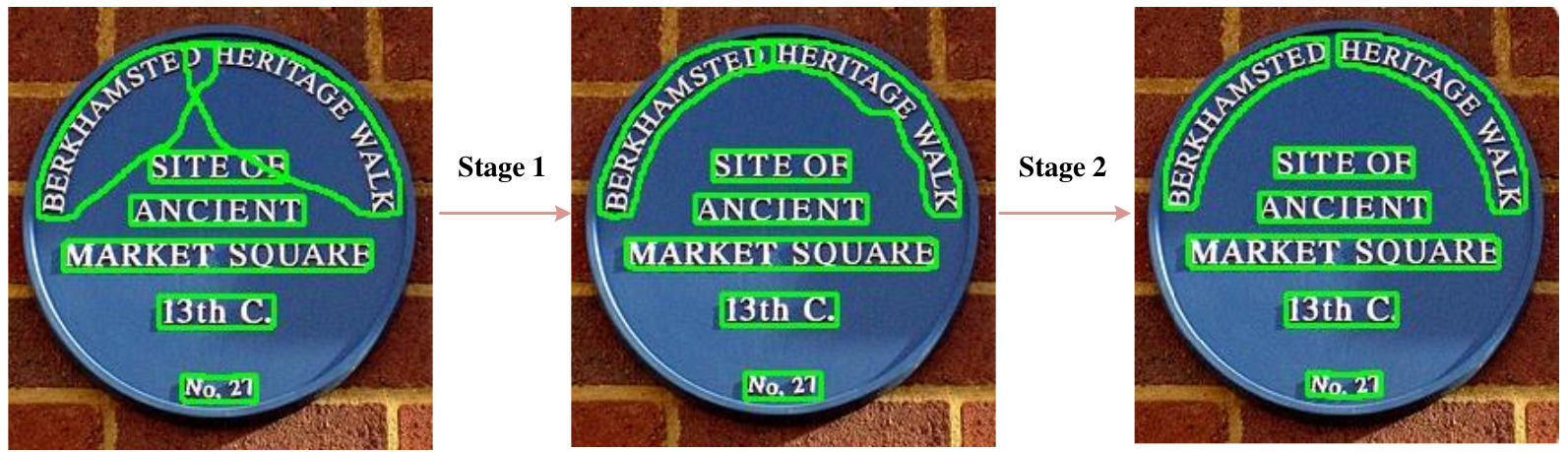}
\caption{\highlight{Visual results of different stages. The green contours are detection boundaries.}
}
\label{fig-stage}
\end{figure}

\begin{table}[]
\setlength\tabcolsep{19.5pt}
\centering
\caption{Results of different vertex numbers in contour transformers on CTW1500~\cite{liu2019curved}.}
\label{tabel-ablation-3}
\begin{tabular}{cccc}
\hline
\multicolumn{1}{c|}{Number} & R                & P               & F               \\ \hline
\multicolumn{1}{c|}{16} & 81.9                 & 87.4                 & 84.6                 \\
\multicolumn{1}{c|}{24} & 81.6                 & $\mathbf{88.5}$                 & 84.9                 \\
\multicolumn{1}{c|}{\textbf{32}} & $\mathbf{82.7}$                 & $87.9$                 & $\mathbf{85.2}$                 \\
\multicolumn{1}{c|}{40} & $82.6$                 & 88.0                 & $\mathbf{85.2}$                 \\
\multicolumn{1}{c|}{48} & $82.1$                 & 88.3                 & 85.1                 \\ \hline
\multicolumn{1}{l}{}        & \multicolumn{1}{l}{} & \multicolumn{1}{l}{} & \multicolumn{1}{l}{}
\end{tabular}
\end{table}

\begin{table}[]
\setlength\tabcolsep{10.5pt}
\centering
\caption{\highlight{Results for our CT-Net with different trade-off parameters on CTW1500 \cite{liu2019curved}.}}
\label{tabel-hyper_para}
\begin{tabular}{cccc|ccc}
\hline
$\lambda_{1}$      & $\lambda_{2}$    & $\lambda_{3}$ & $\lambda_{4}$ & R & P & F \\ \hline
1 & 3 & 0.1 & 2 & 82.5   & 87.6      &   85.0        \\
3 & 1 & 0.1 & 2 & 82.3   & \textbf{87.9}      &  85.1         \\
0.3 & 0.3 & 0.1 & 2 & 82.5   & 87.4      & 84.9          \\
1 & 1 & 0.3 & 2 & 81.9   & 87.3      &  84.5         \\
1 & 1 & 0.03 & 2 & 82.0   & 86.4      & 84.2          \\
1 & 1 & 0.1 & 6 & 82.2   & 87.4      &  84.7         \\
1 & 1 & 0.1 & 0.5    & \textbf{82.8}      & 86.9      & 84.8     \\
\textbf{1} & \textbf{1} & \textbf{0.1} & \textbf{2} & 82.7   & \textbf{87.9}      & \textbf{85.2}          \\ \hline
\end{tabular}
\end{table}

\begin{figure*}[]
    \centering
    \setlength{\abovecaptionskip}{0cm}
    \setlength{\belowcaptionskip}{0cm}
    \subfigure[CTW1500]
    {
        \includegraphics[width=\textwidth,height=3cm]{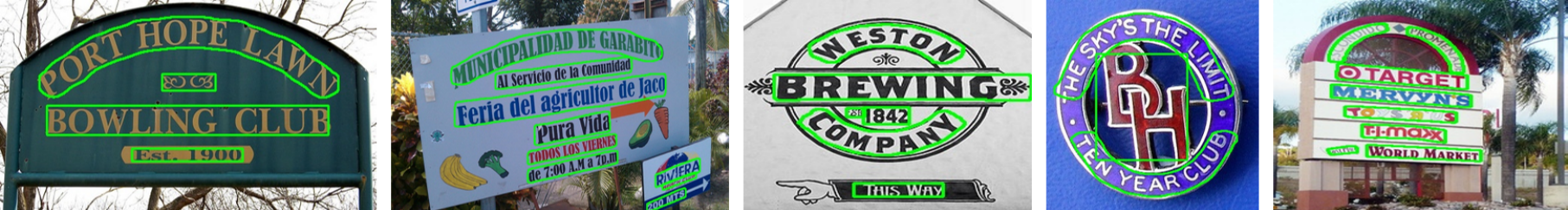}
        \label{fig8a}
    }
    \subfigure[Total-Text]
    {
        \includegraphics[width=\textwidth,height=3cm]{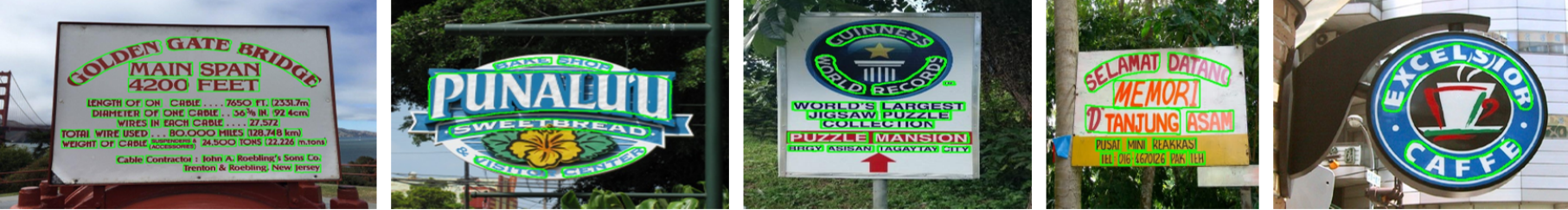}
        \label{fig8b}
    }
    \subfigure[ICDAR2015]
    {
        \includegraphics[width=\textwidth,height=3cm]{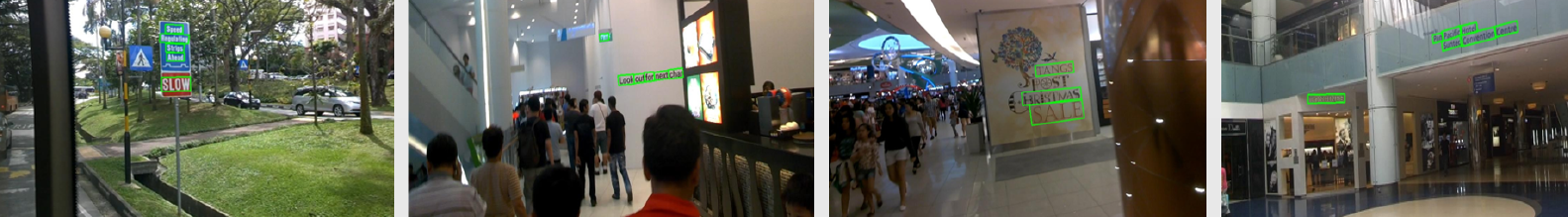}
        \label{fig8c}
    }
    \subfigure[MSRA-TD500]
    {
        \includegraphics[width=\textwidth,height=3cm]{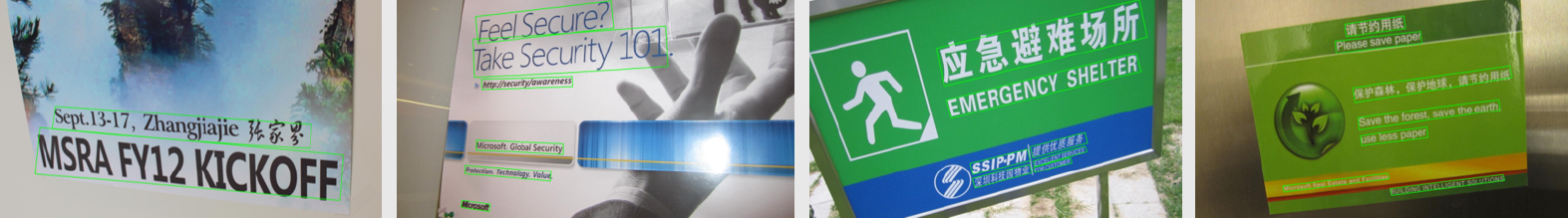}
        \label{fig8d}
    }
    \caption{Example results of our CT-Net on scene text datasets CTW1500~\cite{liu2019curved}, Total-Text~\cite{ch2017total}, ICDAR2015~\cite{karatzas2015icdar}, and MSRA-TD500~\cite{yao2014unified}.}
    \label{fig-visual}

\end{figure*}

\begin{table*}[]

\centering
 \caption{Comparisons with related works on CTW1500~\cite{liu2019curved} and Total-Text~\cite{ch2017total} datasets. Ext means extra training data. Syn, MLT, and ArT denote SynthText\cite{gupta2016synthetic}, ICDAR2017-MLT\cite{nayef2017icdar2017}, and ICDAR-ArT\cite{chng2019icdar2019}, respectively. MixT denotes a dataset composed of COCO-Text\cite{veit2016coco}, SynthCurve\cite{liu2020abcnet}, and ICDAR2019-MLT\cite{nayef2019icdar2019}.}
\label{tabelctwtota}
\begin{tabular}{cc|ccccc|ccccc}

\hline
\multirow{2}{*}{Method} &
\multirow{2}{*}{Paper} &
\multicolumn{5}{c}{CTW1500}                                                                         & \multicolumn{5}{c}{Total-Text}                                                                      \\ \cline{3-12}
                                          &                          &
                                          \multicolumn{1}{c}{Ext} &\multicolumn{1}{c}{Recall} & \multicolumn{1}{c}{Precision} & \multicolumn{1}{c}{F-measure} & FPS  &
                                          \multicolumn{1}{c}{Ext} &\multicolumn{1}{c}{Recall} & \multicolumn{1}{c}{Precision} & \multicolumn{1}{c}{F-measure} &  FPS    \\ \hline
TextSnake\cite{long2018textsnake}               & ECCV'$18$                           & \multicolumn{1}{c}{Syn}                                              & \multicolumn{1}{c}{$\mathbf{85.3}$}   & \multicolumn{1}{c}{$67.9$}      & \multicolumn{1}{c}{$75.6$}      & -
& \multicolumn{1}{c}{Syn}
& \multicolumn{1}{c}{$74.5$}   & \multicolumn{1}{c}{$82.7$}      & \multicolumn{1}{c}{$78.4$}      & -    \\
SegLink++\cite{tang2019seglink++}           & PR'$19$                                                                              &
\multicolumn{1}{c}{Syn}   &
\multicolumn{1}{c}{$79.8$}   & \multicolumn{1}{c}{$82.8$}      & \multicolumn{1}{c}{$81.3$}      & -    &
\multicolumn{1}{c}{Syn}   &
\multicolumn{1}{c}{$80.9$}   & \multicolumn{1}{c}{$82.1$}      & \multicolumn{1}{c}{$81.5$}      & -    \\
TextField\cite{xu2019textfield}         & TIP'$19$                                         & \multicolumn{1}{c}{Syn}                                      & \multicolumn{1}{c}{$79.8$}   & \multicolumn{1}{c}{$83.0$}      & \multicolumn{1}{c}{$81.4$}      & $6.0$    &
\multicolumn{1}{c}{Syn}   &
\multicolumn{1}{c}{$79.9$}   & \multicolumn{1}{c}{$81.2$}      & \multicolumn{1}{c}{$80.6$}      & $6.0$    \\
MSR\cite{2019MSR}              & IJCAI'$19$                                                                          &
\multicolumn{1}{c}{Syn}   &
\multicolumn{1}{c}{$78.3$}   & \multicolumn{1}{c}{$85.0$}      & \multicolumn{1}{c}{$81.5$}      & $4.3$    &
\multicolumn{1}{c}{Syn}   &
\multicolumn{1}{c}{$74.8$}   & \multicolumn{1}{c}{$83.8$}      & \multicolumn{1}{c}{$79.0$}      & $4.3$    \\
ATRR\cite{wang2019arbitrary}              & CVPR'$19$                                                                          &
\multicolumn{1}{c}{-}   &
\multicolumn{1}{c}{$80.2$}   & \multicolumn{1}{c}{$80.1$}      & \multicolumn{1}{c}{$80.1$}      & $10.0$    &
\multicolumn{1}{c}{-}   &
\multicolumn{1}{c}{$76.2$}   & \multicolumn{1}{c}{$80.9$}      & \multicolumn{1}{c}{$78.5$}      & -    \\
CRAFT\cite{baek2019character}              & CVPR'$19$                                                                          &
\multicolumn{1}{c}{Syn}   &
\multicolumn{1}{c}{$81.1$}   & \multicolumn{1}{c}{$86.0$}      & \multicolumn{1}{c}{$83.5$}      & -    &
\multicolumn{1}{c}{Syn}   &
\multicolumn{1}{c}{$79.9$}   & \multicolumn{1}{c}{$87.6$}      & \multicolumn{1}{c}{$83.6$}      & -    \\
PAN\cite{wang2019efficient}              & ICCV'$19$                                                                          &
\multicolumn{1}{c}{Syn}   &
\multicolumn{1}{c}{$81.2$}   & \multicolumn{1}{c}{$86.4$}      & \multicolumn{1}{c}{$83.7$}      & $\mathbf{39.8}$    &
\multicolumn{1}{c}{Syn}   &
\multicolumn{1}{c}{$81.0$}   & \multicolumn{1}{c}{$89.3$}      & \multicolumn{1}{c}{$85.0$}      & $\mathbf{39.6}$    \\
Mask-TTD\cite{liu2019arbitrarily}               & TIP'$20$                                                                         &

\multicolumn{1}{c}{-}   &
\multicolumn{1}{c}{$79.0$}   & \multicolumn{1}{c}{$79.7$}      & \multicolumn{1}{c}{$79.4$}      & -    &
\multicolumn{1}{c}{-}   &
\multicolumn{1}{c}{$74.5$}   & \multicolumn{1}{c}{$79.1$}      & \multicolumn{1}{c}{$76.7$}      & -    \\
TextRay\cite{wang2020textray}              & MM'$20$                                                                         &
\multicolumn{1}{c}{ArT}   &
\multicolumn{1}{c}{$80.4$}   & \multicolumn{1}{c}{$82.8$}      & \multicolumn{1}{c}{$81.6$}      & -    &
\multicolumn{1}{c}{ArT}   &
\multicolumn{1}{c}{$77.9$}   & \multicolumn{1}{c}{$83.5$}      & \multicolumn{1}{c}{$80.6$}      & -    \\
ABCNet\cite{liu2020abcnet}              & CVPR'$20$                                                                        &
\multicolumn{1}{c}{MixT}   &
\multicolumn{1}{c}{$78.5$}   & \multicolumn{1}{c}{$84.4$}      & \multicolumn{1}{c}{$81.4$}      & -    &
\multicolumn{1}{c}{MixT}   &
\multicolumn{1}{c}{$81.3$}   & \multicolumn{1}{c}{$87.9$}      & \multicolumn{1}{c}{$84.5$}      & -    \\
ContourNet\cite{wang2020contournet}              & CVPR'$20$                                                                          &
\multicolumn{1}{c}{-}   &
\multicolumn{1}{c}{$84.1$}   & \multicolumn{1}{c}{$83.7$}      & \multicolumn{1}{c}{$83.9$}      & $4.5$    &
\multicolumn{1}{c}{-}   &
\multicolumn{1}{c}{$83.9$}   & \multicolumn{1}{c}{$86.9$}      & \multicolumn{1}{c}{$85.4$}      & $3.8$    \\
DRRG\cite{2020Deep}              & CVPR'$20$                                                                          &
\multicolumn{1}{c}{MLT}   &
\multicolumn{1}{c}{$83.0$}   & \multicolumn{1}{c}{$85.9$}      & \multicolumn{1}{c}{$84.5$}      & -    &
\multicolumn{1}{c}{MLT}   &
\multicolumn{1}{c}{$\mathbf{84.9}$}   & \multicolumn{1}{c}{$86.5$}      & \multicolumn{1}{c}{$85.7$}      & -    \\
ReLaText\cite{ma2021relatext}              & PR'$21$                                                                          &
\multicolumn{1}{c}{Syn}   &
\multicolumn{1}{c}{$83.3$}   & \multicolumn{1}{c}{$86.2$}      & \multicolumn{1}{c}{$84.8$}      & $10.6$    &
\multicolumn{1}{c}{Syn}   &
\multicolumn{1}{c}{$83.1$}   & \multicolumn{1}{c}{$84.8$}      & \multicolumn{1}{c}{$84.0$}      & $3.2$    \\
OPMP\cite{zhang2020opmp}              & TMM'$21$                                                                       &
\multicolumn{1}{c}{-}   &
\multicolumn{1}{c}{$80.8$}   & \multicolumn{1}{c}{$85.1$}      & \multicolumn{1}{c}{$82.9$}      & $1.4$    &
\multicolumn{1}{c}{Syn}   &
\multicolumn{1}{c}{$82.7$}   & \multicolumn{1}{c}{$87.6$}      & \multicolumn{1}{c}{$85.1$}      & $1.4$    \\
NASK\cite{cao2021all}              & TCSVT'$21$                                                                       &
\multicolumn{1}{c}{Syn}   &
\multicolumn{1}{c}{$80.1$}   & \multicolumn{1}{c}{$83.4$}      & \multicolumn{1}{c}{$81.7$}      & $12.1$    &
\multicolumn{1}{c}{Syn}   &
\multicolumn{1}{c}{$83.2$}   & \multicolumn{1}{c}{$85.6$}      & \multicolumn{1}{c}{$84.4$}      & $8.4$    \\
PCR\cite{dai2021progressive}              & CVPR'$21$                                                                          &
\multicolumn{1}{c}{MLT}   &
\multicolumn{1}{c}{$82.3$}   & \multicolumn{1}{c}{$87.2$}      & \multicolumn{1}{c}{$84.7$}      & -    &
\multicolumn{1}{c}{MLT}   &
\multicolumn{1}{c}{$82.0$}   & \multicolumn{1}{c}{$88.5$}      & \multicolumn{1}{c}{$85.2$}      & -    \\
FCENet\cite{zhu2021fourier}              & CVPR'$21$                                                                          &
\multicolumn{1}{c}{-}   &
\multicolumn{1}{c}{$83.4$}   & \multicolumn{1}{c}{$87.6$}      & \multicolumn{1}{c}{$\mathbf{85.5}$}      & -    &
\multicolumn{1}{c}{-}   &
\multicolumn{1}{c}{$82.5$}   & \multicolumn{1}{c}{$89.3$}      & \multicolumn{1}{c}{$85.8$}      & -    \\
TextBPN\cite{zhang2021adaptive}              & ICCV'$21$                                                                          &
\multicolumn{1}{c}{Syn}   &
\multicolumn{1}{c}{$81.4$}   & \multicolumn{1}{c}{$87.8$}      & \multicolumn{1}{c}{$84.5$}      & $12.1$    &
\multicolumn{1}{c}{Syn}   &
\multicolumn{1}{c}{$84.6$}   & \multicolumn{1}{c}{$90.2$}      & \multicolumn{1}{c}{$87.3$}      & 12.6    \\
DText \cite{cai2022arbitrarily}                    & PR'$22$                                                                       &
\multicolumn{1}{c}{Syn}   &
\multicolumn{1}{c}{$\mathbf{82.7}$}   & \multicolumn{1}{c}{$86.9$}      & \multicolumn{1}{c}{$84.7$}      & $-$ &
\multicolumn{1}{c}{Syn}   &
\multicolumn{1}{c}{$82.7$}   & \multicolumn{1}{c}{$90.5$}      & \multicolumn{1}{c}{$86.4$}      & $-$ \\
TextDCT \cite{su2022textdct}                    & TMM'$22$                                                                       &
\multicolumn{1}{c}{Syn}   &
\multicolumn{1}{c}{$\mathbf{85.3}$}   & \multicolumn{1}{c}{$85.0$}      & \multicolumn{1}{c}{$85.1$}      & $17.2$ &
\multicolumn{1}{c}{Syn}   &
\multicolumn{1}{c}{$82.7$}   & \multicolumn{1}{c}{$87.2$}      & \multicolumn{1}{c}{$84.9$}      & $15.1$ \\
Zhao \etal\cite{zhao2022mixed}                    & TIP'$22$                                                                       &
\multicolumn{1}{c}{Syn}   &
\multicolumn{1}{c}{$82.1$}   & \multicolumn{1}{c}{$86.1$}      & \multicolumn{1}{c}{$84.1$}      & - &
\multicolumn{1}{c}{Syn}   &
\multicolumn{1}{c}{$83.3$}   & \multicolumn{1}{c}{$88.2$}      & \multicolumn{1}{c}{$85.6$}      & - \\
DBNet++\cite{liao2022real}              & TPAMI'$22$                                                                          &
\multicolumn{1}{c}{Syn}   &
\multicolumn{1}{c}{$82.8$}   & \multicolumn{1}{c}{$87.9$}      & \multicolumn{1}{c}{$85.3$}      & $26.0$    &
\multicolumn{1}{c}{Syn}   &
\multicolumn{1}{c}{$83.2$}   & \multicolumn{1}{c}{$88.9$}      & \multicolumn{1}{c}{$86.0$}      & $28.0$    \\
Tang \etal\cite{tang2022few}              & CVPR'$22$                                                                          &
\multicolumn{1}{c}{Syn}   &
\multicolumn{1}{c}{$82.4$}   & \multicolumn{1}{c}{$88.1$}      & \multicolumn{1}{c}{$85.2$}      & $-$    &
\multicolumn{1}{c}{Syn}   &
\multicolumn{1}{c}{$85.7$}   & \multicolumn{1}{c}{$90.7$}      & \multicolumn{1}{c}{$\mathbf{88.1}$}      & $-$    \\
\hline
\textbf{CT-Net}
                & Ours                                                                          &
\multicolumn{1}{c}{-}   &
\multicolumn{1}{c}{$82.7$}   & \multicolumn{1}{c}{$87.9$}      & \multicolumn{1}{c}{$85.2$}      & $11.3$ &
\multicolumn{1}{c}{-}   &
\multicolumn{1}{c}{$83.6$}   & \multicolumn{1}{c}{$89.2$}      & \multicolumn{1}{c}{$86.3$}      & $9.8$ \\
\textbf{CT-Net}                    & Ours                                                                       &
\multicolumn{1}{c}{Syn}   &
\multicolumn{1}{c}{$83.8$}   & \multicolumn{1}{c}{$\mathbf{88.5}$}      & \multicolumn{1}{c}{$\mathbf{86.1}$}      & $11.2$ &
\multicolumn{1}{c}{Syn}   &
\multicolumn{1}{c}{$85.0$}   & \multicolumn{1}{c}{$\mathbf{90.8}$}      & \multicolumn{1}{c}{$87.8$}      & $10.1$ \\ \hline

\end{tabular}
\end{table*}

As can be seen in
Table~\ref{tabel2}, our contour transformer achieves the best performance compared with the other two methods on both CTW1500 and Total-Text. It achieves improvements on CTW1500 by 0.6 F-measure over CCN and 0.5 F-measure over GCN, respectively. This is because our contour refinement module can obtain the global contour information more effectively. \highlight{Besides, we compare the number of parameters (Params), floating point operations (FLOPs), and FPS of these three structures. Compared to CCN and GCN, our CT has less parameters and computational complexity, and thus has faster running speed (0.7 FPS over CCN and 0.4 FPS over GCN on CTW1500).}
%so equipped with CT

\subsubsection{Adaptive Training Strategy}

Here we investigate the effectiveness of adaptive training strategy. The results on CTW1500 dataset are shown in Table~\ref{tabel-ablation-1}. The F-measure is increased by $0.5$ after employing the adaptive training strategy. This gain can be ascribed that the adaptive training strategy enables the contour transformer to learn more potential deformation paths.

\subsubsection{Re-score Mechanism}

Table~\ref{tabel-ablation-1} shows the results of using the re-score mechanism over BCNet and BCANet on CTW1500 dataset. We can see that BCRNet and CT-Net both achieve the gain of $0.7$ F-measure after using the re-score mechanism. This proves that the re-score mechanism can effectively evaluate the confidence of the contours.

\subsubsection{\highlight{Stage number}}

To explore the influence of \highlight{stage number}, we compare our CT-Net with different stages. As shown in Table~\ref{tabel-ablation-2}, with the increase of the number of \highlight{stages}, the detection performance is gradually improved to a critical value, but the inference speed is gradually dropped. As the stage increases from $1$ to $2$, the F-measure boosts from $84.8$ to $85.2$. After further increasing to $3$, the result is not improved while the inference time is decreased from $12.2$ FPS to $11.2$ FPS. \highlight{The example results across stages are shown in Fig.~\ref{fig-stage}, which indicate that two contour refinement modules are sufficient to refine an initial contour often with large prediction deviations.} Therefore, considering the balance of speed and performance, we set the number of \highlight{stages} to $2$ in our all experiments.
%As shown in Fig. 7, the detection boundaries become more accurate along with the increase of iterations.

\subsubsection{Contour Vertex Number}

In the contour refinement module, the number of input vertexes $N_a$ for each text affects the fineness of the predicted contour. As shown in Table~\ref{tabel-ablation-3}, CT-Net improves by $0.6$ in F-measure when $N_a$ is increased from $16$ to $32$.
% When $N_a$ is further increased, .
With further increase of $N_a$, the change of detection performance is insignificant. This demonstrates that our contour refinement module can effectively capture the long-range context dependencies between contour vertexes.

\subsubsection{\highlight{Trade-Off Parameter}}

\highlight{To investigate the influence of different trade-off parameter values in the full loss function on our method, we implement CT-Net with different values of $\lambda_1$, $\lambda_2$, $\lambda_3$, and $\lambda_4$, as presented in Table~\ref{tabel-hyper_para}. It can be observed that the results are stable for different values, which demonstrates the robustness of our method on the values of trade-off parameters. Note that the impacts of changes to $\lambda_3$ and $\lambda_4$ are slightly greater than those of changes to $\lambda_1$ and $\lambda_2$. This is because $\lambda_3$ and $\lambda_4$ are related to the joint optimization of initial contour prediction and contour refinement, and thus brings greater impacts on the whole model.}
%modules are jointly optimized, the update to the latter module

\begin{table*}[]
\centering
 \caption{Comparisons with state-of-the-art works on ICDAR2015~\cite{karatzas2015icdar} and MSRA-TD500~\cite{yao2014unified}. CO-T denotes COCO-Text\cite{veit2016coco}.}
 \label{tabelic157}
\centering
\begin{tabular}{cc|ccccc|ccccc}
\hline
\multirow{2}{*}{Method} &
\multirow{2}{*}{Paper}&
\multicolumn{5}{c}{ICDAR2015}                                                    & \multicolumn{5}{c}{MSRA-TD500}                                                                      \\ \cline{3-12}
                          &                    &  \multicolumn{1}{c}{Ext} &
                      \multicolumn{1}{c}{Recall} &  \multicolumn{1}{c}{Precision} &  \multicolumn{1}{c}{F-measure}  &
                     FPS  &
                 \multicolumn{1}{c}{Ext} &
                     \multicolumn{1}{c}{Recall}  & \multicolumn{1}{c}{Precision} & \multicolumn{1}{c}{F-measure} & FPS      \\ \hline

RRPN\cite{ma2018arbitrary}                 & TMM'$18$                                                                         & \multicolumn{1}{c}{-}   & \multicolumn{1}{c}{$73.0$}      &
\multicolumn{1}{c}{$82.0$}      &
\multicolumn{1}{c}{$77.0$}      & -
& \multicolumn{1}{c}{-} & \multicolumn{1}{c}{$68.0$}   & \multicolumn{1}{c}{$82.0$}      & \multicolumn{1}{c}{$74.4$}      & 4.4 \\
Cheng \etal\cite{cheng2019direct}                 & TCSVT'$19$                                                                         & \multicolumn{1}{c}{CO-T}   & \multicolumn{1}{c}{$82.0$}      &
\multicolumn{1}{c}{$90.0$}      &
\multicolumn{1}{c}{$86.0$}      & $8.6$
& \multicolumn{1}{c}{-} & \multicolumn{1}{c}{-}   & \multicolumn{1}{c}{-}      & \multicolumn{1}{c}{-}      & - \\
ATRR\cite{wang2019arbitrary}              & CVPR'$19$                                                                          &
\multicolumn{1}{c}{-}   &
\multicolumn{1}{c}{$86.0$}   & \multicolumn{1}{c}{$89.2$}      & \multicolumn{1}{c}{$87.6$}      & -    &
\multicolumn{1}{c}{-}   &
\multicolumn{1}{c}{82.1}   & \multicolumn{1}{c}{85.2}      & \multicolumn{1}{c}{83.6}      & -    \\
CRAFT\cite{baek2019character}                 & CVPR'$19$                                                                         & \multicolumn{1}{c}{Syn}   & \multicolumn{1}{c}{$84.3$}      &
\multicolumn{1}{c}{$89.8$}      &
\multicolumn{1}{c}{$86.9$}      & $8.6$
& \multicolumn{1}{c}{Syn} & \multicolumn{1}{c}{$78.2$}   & \multicolumn{1}{c}{$88.2$}      & \multicolumn{1}{c}{$82.9$}      & $8.6$ \\
DRRG\cite{2020Deep}                 & CVPR'$20$                                                                         & \multicolumn{1}{c}{MLT}   & \multicolumn{1}{c}{$84.7$}      &
\multicolumn{1}{c}{$88.5$}      &
\multicolumn{1}{c}{$86.6$}      & -
& \multicolumn{1}{c}{MLT} & \multicolumn{1}{c}{$82.3$}   & \multicolumn{1}{c}{$80.5$}      & \multicolumn{1}{c}{$85.1$}      & - \\
ContourNet\cite{wang2020contournet}              & CVPR'$20$                                                                        & \multicolumn{1}{c}{-}   & \multicolumn{1}{c}{$86.1$}      &
\multicolumn{1}{c}{$87.6$}      &
\multicolumn{1}{c}{$86.9$}      & $3.5$
& \multicolumn{1}{c}{-} & \multicolumn{1}{c}{-}   & \multicolumn{1}{c}{-}      & \multicolumn{1}{c}{-}      & - \\
R-Net\cite{wang2020r}                 & TMM'$21$                                                                         & \multicolumn{1}{c}{Syn}   & \multicolumn{1}{c}{$82.8$}      &
\multicolumn{1}{c}{$88.7$}      &
\multicolumn{1}{c}{$85.6$}      & $\mathbf{21.4}$
& \multicolumn{1}{c}{Syn} & \multicolumn{1}{c}{$79.7$}   & \multicolumn{1}{c}{$83.7$}      & \multicolumn{1}{c}{$81.7$}      & $11.8$ \\
FCENet\cite{zhu2021fourier}                 & CVPR'$21$                                                                         & \multicolumn{1}{c}{-}   & \multicolumn{1}{c}{$82.6$}      &
\multicolumn{1}{c}{$90.1$}      &
\multicolumn{1}{c}{$86.2$}      & -
& \multicolumn{1}{c}{-} & \multicolumn{1}{c}{-}   & \multicolumn{1}{c}{-}      & \multicolumn{1}{c}{-}      & - \\

MOST\cite{he2021most}                 & CVPR'$21$                                                                         & \multicolumn{1}{c}{Syn}   & \multicolumn{1}{c}{$\mathbf{87.3}$}      &
\multicolumn{1}{c}{$89.1$}      &
\multicolumn{1}{c}{$88.2$}      & $10.0$
& \multicolumn{1}{c}{Syn} & \multicolumn{1}{c}{$82.7$}   & \multicolumn{1}{c}{$90.4$}      & \multicolumn{1}{c}{$86.4$}      & $\mathbf{51.8}$ \\

PCR\cite{dai2021progressive}              & CVPR'$21$                                                                          &
\multicolumn{1}{c}{-}   &
\multicolumn{1}{c}{-}   & \multicolumn{1}{c}{-}      & \multicolumn{1}{c}{-}      & -    &
\multicolumn{1}{c}{MLT}   &
\multicolumn{1}{c}{$83.5$}   & \multicolumn{1}{c}{$90.8$}      & \multicolumn{1}{c}{$87.0$}      & -    \\
TextBPN\cite{zhang2021adaptive}              & ICCV'$21$                                                                          &
\multicolumn{1}{c}{-}   &
\multicolumn{1}{c}{-}   & \multicolumn{1}{c}{-}      & \multicolumn{1}{c}{-}      & -    &
\multicolumn{1}{c}{Syn}   &
\multicolumn{1}{c}{$80.7$}   & \multicolumn{1}{c}{$85.4$}      & \multicolumn{1}{c}{$83.0$}      & 12.7\\
DText \cite{cai2022arbitrarily}                    & PR'$22$                                                                       &
\multicolumn{1}{c}{Syn}   &
\multicolumn{1}{c}{$85.6$}   & \multicolumn{1}{c}{$88.5$}      & \multicolumn{1}{c}{$87.0$}      & - &
\multicolumn{1}{c}{Syn}   &
\multicolumn{1}{c}{$83.1$}   & \multicolumn{1}{c}{$87.9$}      & \multicolumn{1}{c}{$85.4$}      & - \\
TextDCT \cite{su2022textdct}                 & TMM'$22$                                                                         & \multicolumn{1}{c}{Syn}   & \multicolumn{1}{c}{$84.8$}      &
\multicolumn{1}{c}{$88.9$}      &
\multicolumn{1}{c}{$86.8$}      & $7.5$
& \multicolumn{1}{c}{-} & \multicolumn{1}{c}{-}   & \multicolumn{1}{c}{-}      & \multicolumn{1}{c}{-}      & - \\
Zhao \etal\cite{zhao2022mixed}                    & TIP'$22$                                                                       &
\multicolumn{1}{c}{Syn}   &
\multicolumn{1}{c}{$82.4$}   & \multicolumn{1}{c}{$89.4$}      & \multicolumn{1}{c}{$85.8$}      & - &
\multicolumn{1}{c}{Syn}   &
\multicolumn{1}{c}{$81.1$}   & \multicolumn{1}{c}{$88.7$}      & \multicolumn{1}{c}{$84.7$}      & - \\
DBNet++\cite{liao2022real}              & TPAMI'$22$                                                                          &
\multicolumn{1}{c}{Syn}   &
\multicolumn{1}{c}{$83.9$}   & \multicolumn{1}{c}{$\mathbf{90.9}$}      & \multicolumn{1}{c}{$87.3$}      & $10.0$    &
\multicolumn{1}{c}{Syn}   &
\multicolumn{1}{c}{$83.3$}   & \multicolumn{1}{c}{$\mathbf{91.5}$}      & \multicolumn{1}{c}{$87.2$}      & $29.0$    \\
RFN\cite{guan2022industrial}              & TCSVT'$22$                                                                          &
\multicolumn{1}{c}{-}   &
\multicolumn{1}{c}{-}   & \multicolumn{1}{c}{-}      & \multicolumn{1}{c}{-}      & -    &
\multicolumn{1}{c}{-}   &
\multicolumn{1}{c}{$80.0$}   & \multicolumn{1}{c}{$88.4$}      & \multicolumn{1}{c}{$84.0$}      & -    \\
Keserwani \etal\cite{keserwani2022robust}              & TCSVT'$22$                                                                          &
\multicolumn{1}{c}{-}   &
\multicolumn{1}{c}{$82.7$}   & \multicolumn{1}{c}{$85.9$}      & \multicolumn{1}{c}{$84.3$}      & -    &
\multicolumn{1}{c}{-}   &
\multicolumn{1}{c}{-}   & \multicolumn{1}{c}{-}      & \multicolumn{1}{c}{-}      & -    \\
\hline

\textbf{CT-Net}                 & Ours                                                                         & \multicolumn{1}{c}{-}   & \multicolumn{1}{c}{$85.6$}      &
\multicolumn{1}{c}{$88.1$}      &
\multicolumn{1}{c}{$86.8$}      & $6.6$
& \multicolumn{1}{c}{-} & \multicolumn{1}{c}{$80.4$}   & \multicolumn{1}{c}{$89.8$}      & \multicolumn{1}{c}{$84.8$}      & $11.4$ \\

\textbf{CT-Net}                 & Ours                                                                         & \multicolumn{1}{c}{Syn}   & \multicolumn{1}{c}{$86.4$}      &
\multicolumn{1}{c}{$\mathbf{90.9}$}      &
\multicolumn{1}{c}{$\mathbf{88.6}$}      & $6.5$
& \multicolumn{1}{c}{Syn} & \multicolumn{1}{c}{$\mathbf{84.4}$}   & \multicolumn{1}{c}{$90.8$}      & \multicolumn{1}{c}{$\mathbf{87.5}$}      & $11.6$ \\
\hline

\end{tabular}
\end{table*}

\subsection{Comparison with State-of-the-Art Methods}

We compare our CT-Net with previous works on four benchmarks, including two benchmarks for multi-oriented texts and two benchmarks for curved texts. Some qualitative results are visualized in Fig. \ref{fig-visual}, which demonstrates the effectiveness of our CT-Net on scene texts with variety and complexity in size, font, and orientation.

\subsubsection{Evaluation on Long Curved Text Benchmark}

The comparison results on long curved text dataset CTW1500 are given in Table~\ref{tabelctwtota}, from which we can see that CT-Net achieves the state-of-the-art results of $83.8$, $88.5$ and $86.1$ in recall, precision and F-measure, respectively.
Compared with the most recent transformer based work \cite{tang2022few}, %which also uses transformer to locate texts,
CT-Net achieves a $0.9$ improvement in F-measure with a more concise structure. Besides, our progressive contour regression method soundly outperforms multi-stage contour based methods \cite{dai2021progressive,zhang2021adaptive}. \highlight{This can be attributed to two main reasons. First, our proposed adaptive training strategy with a re-score mechanism can reduce error accumulation across stages. Second, our contour transformer module is more efficiently to extract global contour information, which is beneficial for perceiving the optimized direction of contour points. As shown in Fig.~\ref{fig-stage}, although there is a large deviation in the prediction of the initial contour, our contour refinement module still can accurately correct it.}

\subsubsection{Evaluation on Curved Text Benchmark}

To test the model performance for detecting curved texts, we compare our CT-Net with other works on Total-Text dataset. The results on Total-Text are shown in Table~\ref{tabelctwtota}, where we evaluate CT-Net using the protocol in~\cite{ch2017total}. We can observe that our CT-Net achieves competitive results and speed. Specifically, CT-Net is superior to previous regression based methods such as ATRR \cite{wang2019arbitrary}, ContourNet \cite{wang2020contournet}, and TextDCT\cite{su2022textdct}.
Besides, compared to the single-stage contour based approach like TextRay\cite{wang2020textray} and FCENet\cite{zhu2021fourier}, CT-Net achieves more accurate localization through multiple perceptions. For example, CT-Net significantly boosts the recall of $2.5$, precision of $1.5$, and F-measure of $2.0$, compared with the state-of-the-art single-stage contour based method FCENet. \highlight{This is mainly because for some complex texts, such single-stage method is difficult to accurately obtain their overall layout with only a single perception. In contrast, our CT-Net can iteratively correct the deviations by multiple perceptions on the text contours, as shown in Fig.~\ref{fig-stage}.}
%caused by a single perception through

\subsubsection{Evaluation on Oriented Text Benchmark}

We evaluate our CT-Net on ICDAR2015 dataset to test its effectiveness of arbitrary-orientated text detection. ICDAR2015 is one of the most popular multi-oriented text datasets with mostly low text quality and complex backgrounds.
Test results on ICDAR2015 following the standard evaluation metric are shown in Table~\ref{tabelic157}. We can see that our CT-Net achieves $85.6$, $88.1$, and $86.8$ for recall, precision, and F-measure respectively without any extra datasets, \highlight{and achieves satisfactory results in terms of F-measure ($88.6$ in F-measure) when pre-trained on SynthText. Compared with the performance of the recent segmentation based and connected component based methods \cite{liao2022real,2020Deep} with pre-training, our CT-Net can achieve competitive performance without pre-training, which indicates that our method is more robust to noises by modeling text boundaries from a top-down perspective.}
%comparable to
% that are close to its pre-trained results
% and achieves $88.6$ in F-measure, which significantly outperforms the recent state-of-the-art methods like DBNet++\cite{liao2022real}.

% Table~\ref{tabelic15} presents the results on ICDAR2015 with multi-oriented texts.
% % The quantitative comparisons with other methods on this dataset is listed in .
% We can see that CT-Net achieves the best performance.
% % successfully detects multi-orientated texts.
% Notably, CT-Net achieves $88.4$ in F-measure, which significantly outperforms the recent state-of-the-art methods like FCENet.

\subsubsection{Evaluation on Multi-Lingual Benchmark}

The comparison results on MSRA-TD500 dataset for evaluating the ability of multi-lingual text detection are presented in Table~\ref{tabelic157}. CT-Net outperforms at least $0.5$ than multi-stage contour based methods \cite{dai2021progressive,zhang2021adaptive} in F-measure. Besides, \highlight{due to the small number of training images in MSRA-TD500 dataset, the previous methods \cite{liao2022real,2020Deep,wang2019arbitrary} usually require pre-training on other datasets to achieve better results. In contrast, our method can achieve comparable performance without pre-training, which demonstrates the good generalization ability of our method.} The example results on MSRA-TD500 are shown in Fig.~\ref{fig-visual}, which shows that our CT-Net can effectively detect multi-lingual and long texts.

\begin{figure}
    \centering
    \subfigure[]
    {
        \includegraphics[width=0.22\textwidth,height=2.8cm]{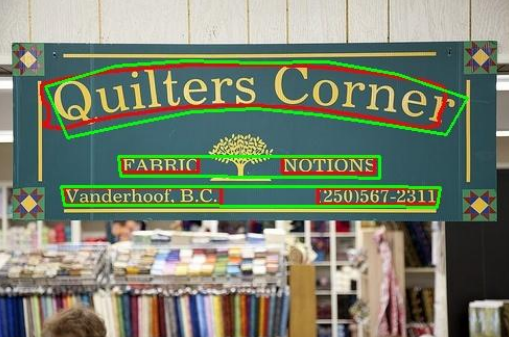}
        \label{fig-limitation1}
    }
    \subfigure[]
    {
        \includegraphics[width=0.22\textwidth,height=2.8cm]{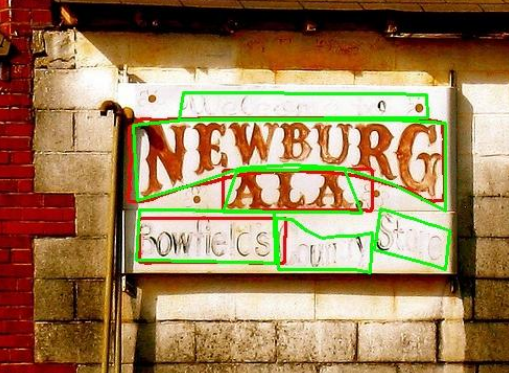}
        \label{fig-limitation2}
    }
    \caption{Failure cases, in which green contours are ground-truths while red contours are predicted results. (a) Unreasonable annotations. (b) Challenging illuminations.}
    \label{fig-limitation}
\end{figure}

\subsection{Limitations}
According to the above experimental results, our CT-Net can perform well in most challenging scenes. However, there are a few failure cases as shown in Fig.~\ref{fig-limitation}. Although our CT-Net can alleviate false positives through adaptive training strategy and re-score mechanism, it cannot solve false positives caused by unreasonable annotations. Moreover, CT-Net cannot work well under challenging illuminations like over- and under-exposures, %struggles with over-exposure or badly illuminated text regions,
owing to the unclear text boundaries.
% \highlight{it
% cannot balance efficiency and accuracy because the multi-stage optimized pipeline in CT-Net is complex and require larger input image sizes to detect small-sized texts.}
% Furthermore,
\highlight{On the other hand, it is hard for our method to balance efficiency and accuracy, since the multi-stage optimized pipeline in CT-Net is complex and requires larger input image sizes to detect small-sized texts. We will try to improve the robustness on challenging annotations and illuminations, and explore a more powerful structure in the future work.}

\section{Conclusion}
In this paper, we have proposed a novel multi-stage contour based framework for arbitrary-shaped scene text detection. First, we adopt a contour initialization module to generate initial contours. Then, we use multiple contour refinement modules with an adaptive training strategy and a re-score mechanism to perform iterative contour refinement, which is beneficial for obtaining more accurate text shapes. Extensive experiments have shown that our method can precisely detect arbitrary-shaped texts in challenging benchmark datasets. In the future study, we will develop this work and explore an end-to-end scene text spotting framework.

\bibliographystyle{IEEEtran}
\bibliography{IEEEabrv,reference}

\begin{IEEEbiography}[{\includegraphics[width=1in,height=1.25in,clip,keepaspectratio]{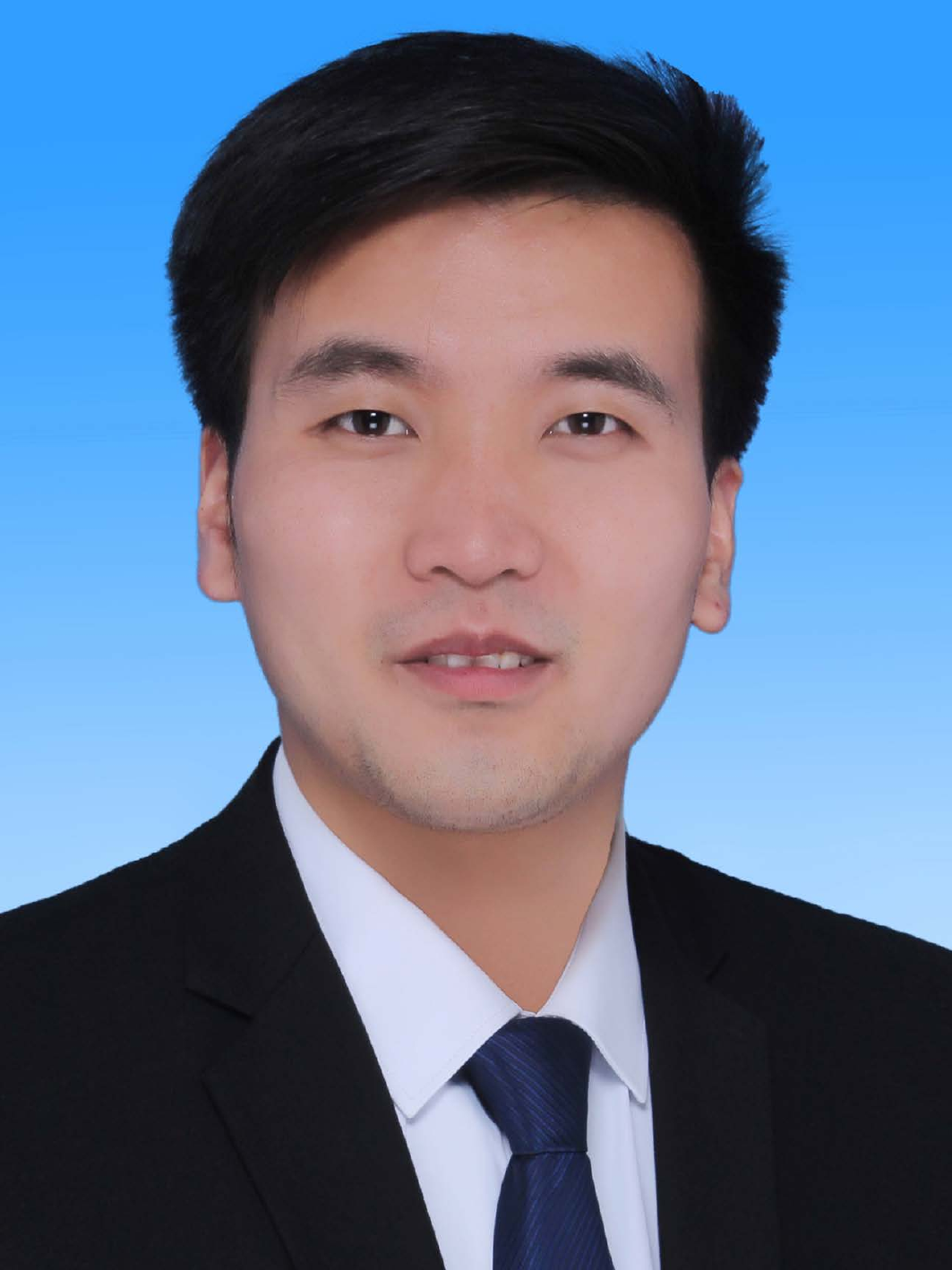}}]{Zhiwen Shao}
received his B.Eng. degree in Computer Science and Technology from the Northwestern Polytechnical University, China in 2015. He received the Ph.D. degree from the Shanghai Jiao Tong University, China in 2020. He is now an Associate Professor at the School of Computer Science and Technology, China University of Mining and Technology, China. From 2017 to 2018, he was a joint Ph.D. student at the Multimedia and Interactive Computing Lab, Nanyang Technological University, Singapore. His research interests lie in computer vision and deep learning. He has been serving as a PC member in IJCAI and AAAI.
\end{IEEEbiography}

\begin{IEEEbiography}[{\includegraphics[width=1in,height=1.25in,clip,keepaspectratio]{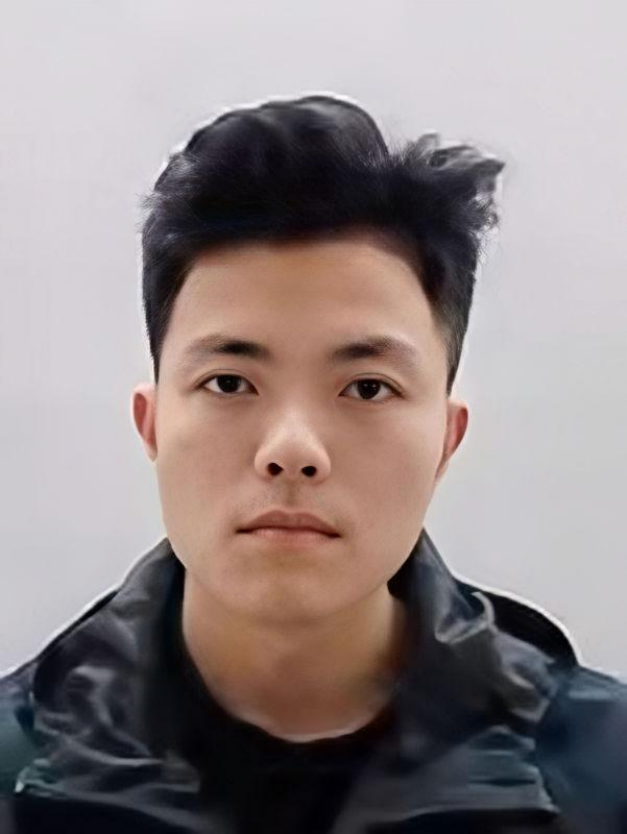}}]{Yuchen Su} is currently pursuing the M.S. degree at the School of Computer Science and Technology, China University of Mining and Technology, China, under the supervision of Prof. Yong Zhou, Prof. Fanrong Meng, and Dr. Zhiwen Shao. His current research interests include scene text detection and recognition.
\end{IEEEbiography}

\begin{IEEEbiography}[{\includegraphics[width=1in,height=1.25in,clip,keepaspectratio]{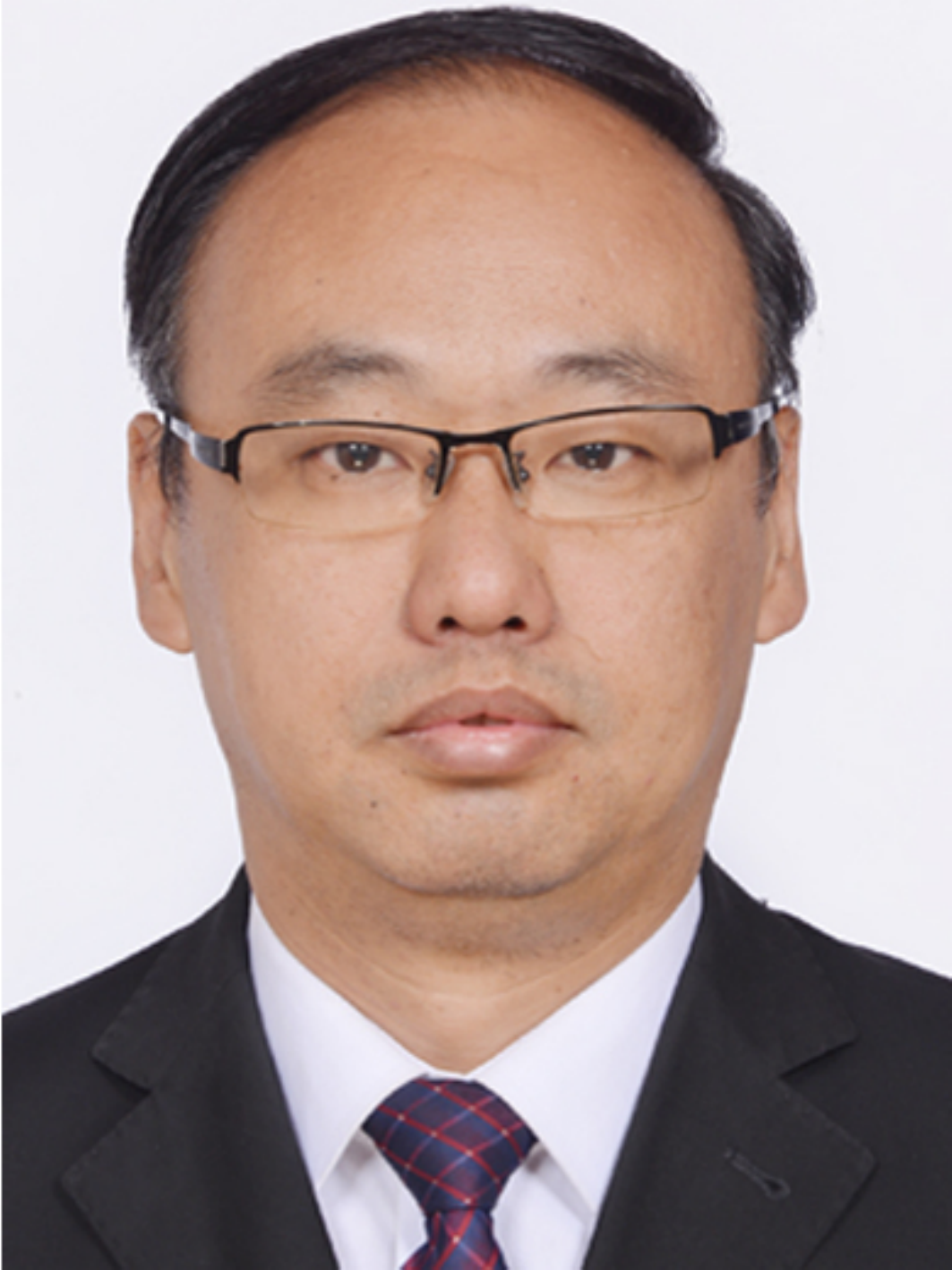}}]{Yong Zhou} received the M.S. and Ph.D. degrees in Control Theory and Control Engineering from the China University of Mining and Technology, China in 2003 and 2006, respectively. He is currently a Professor with the School of Computer Science and Technology, China University of Mining and Technology, China. His research interests include machine learning, intelligence
optimization, and data mining. He has been serving as an Associate Editor of ACM TOMM.
\end{IEEEbiography}

\begin{IEEEbiography}[{\includegraphics[width=1in,height=1.25in,clip,keepaspectratio]{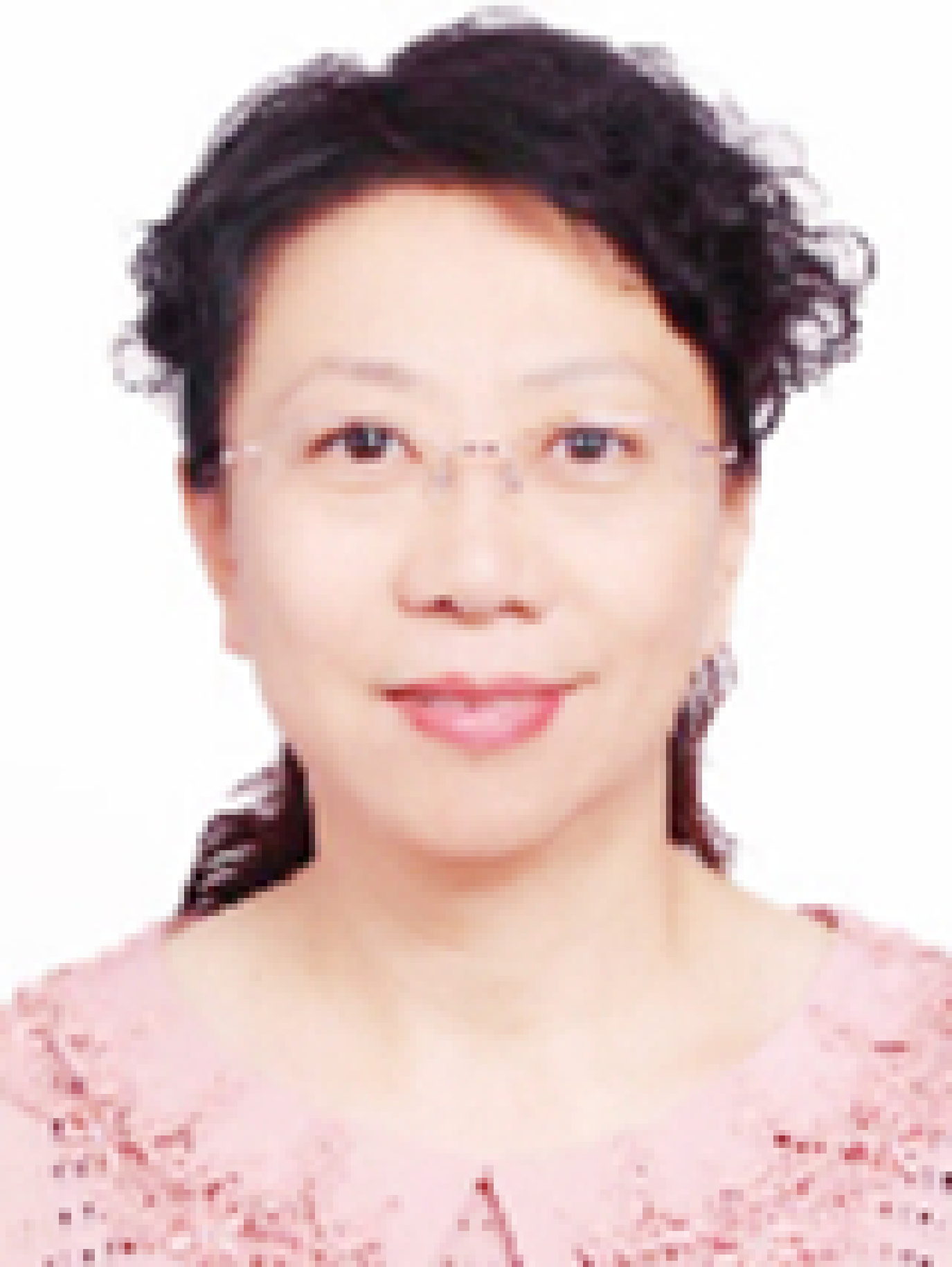}}]{Fanrong Meng} received the Ph.D. degree from the China University of Mining and Technology, China. She is currently a Professor with the School of Computer Science and Technology, China University of Mining and Technology, China. Her research interests include intelligent information processing, database technology, and data mining.
\end{IEEEbiography}

\begin{IEEEbiography}[{\includegraphics[width=1in,height=1.25in,clip,keepaspectratio]{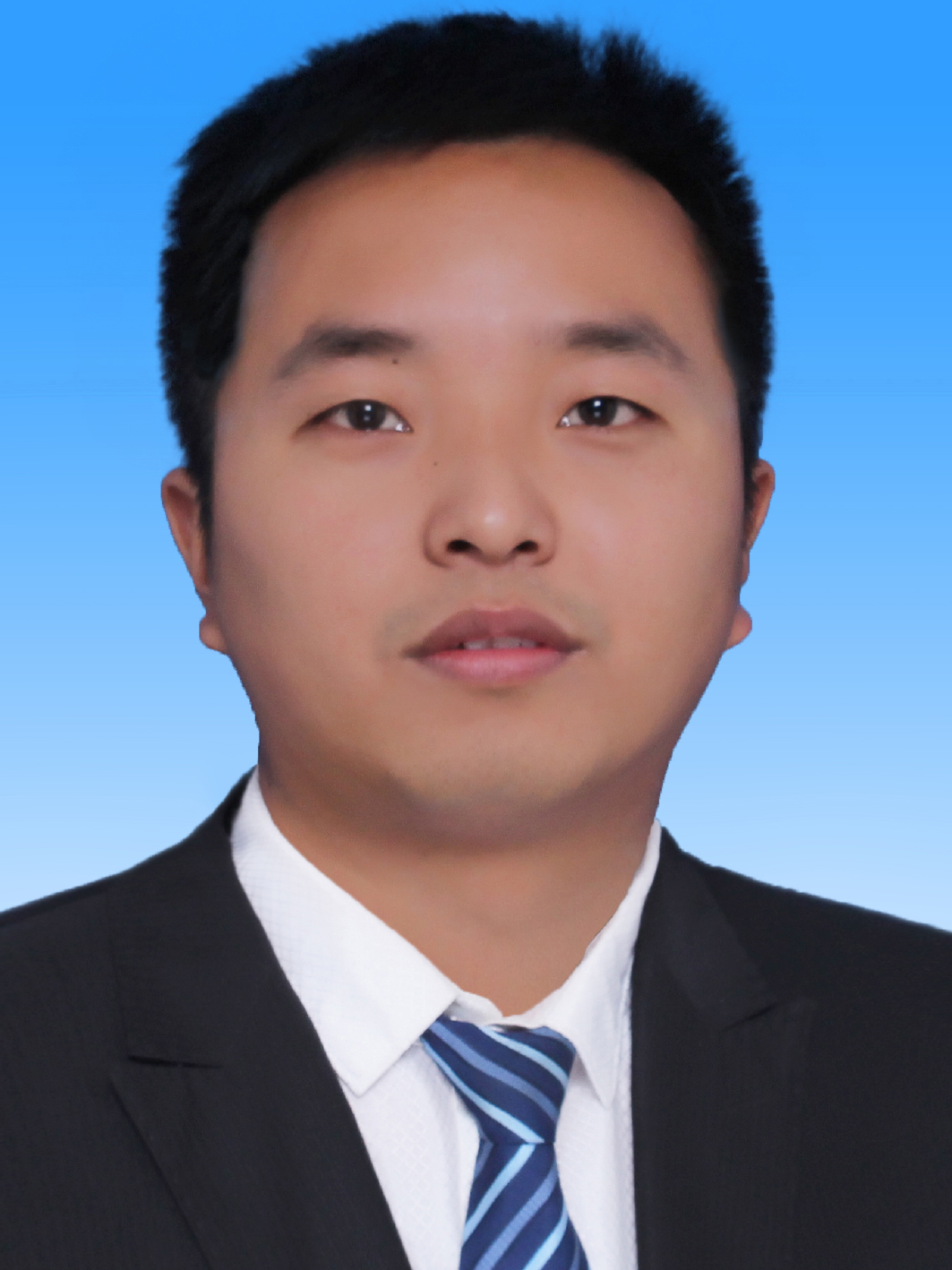}}]{Hancheng Zhu} received the B.S. degree from the Changzhou Institute of Technology, Changzhou, China, in 2012, and the M.S. and Ph.D. degrees from the China University of Mining and Technology, Xuzhou, China, in 2015 and 2020, respectively. He is currently a Tenure-Track Associate Professor at the School of Computer Science and Technology, China University of Mining and Technology, China. His research interests include image aesthetics assessment and affective computing.
\end{IEEEbiography}

\begin{IEEEbiography}[{\includegraphics[width=1in,height=1.25in,clip,keepaspectratio]{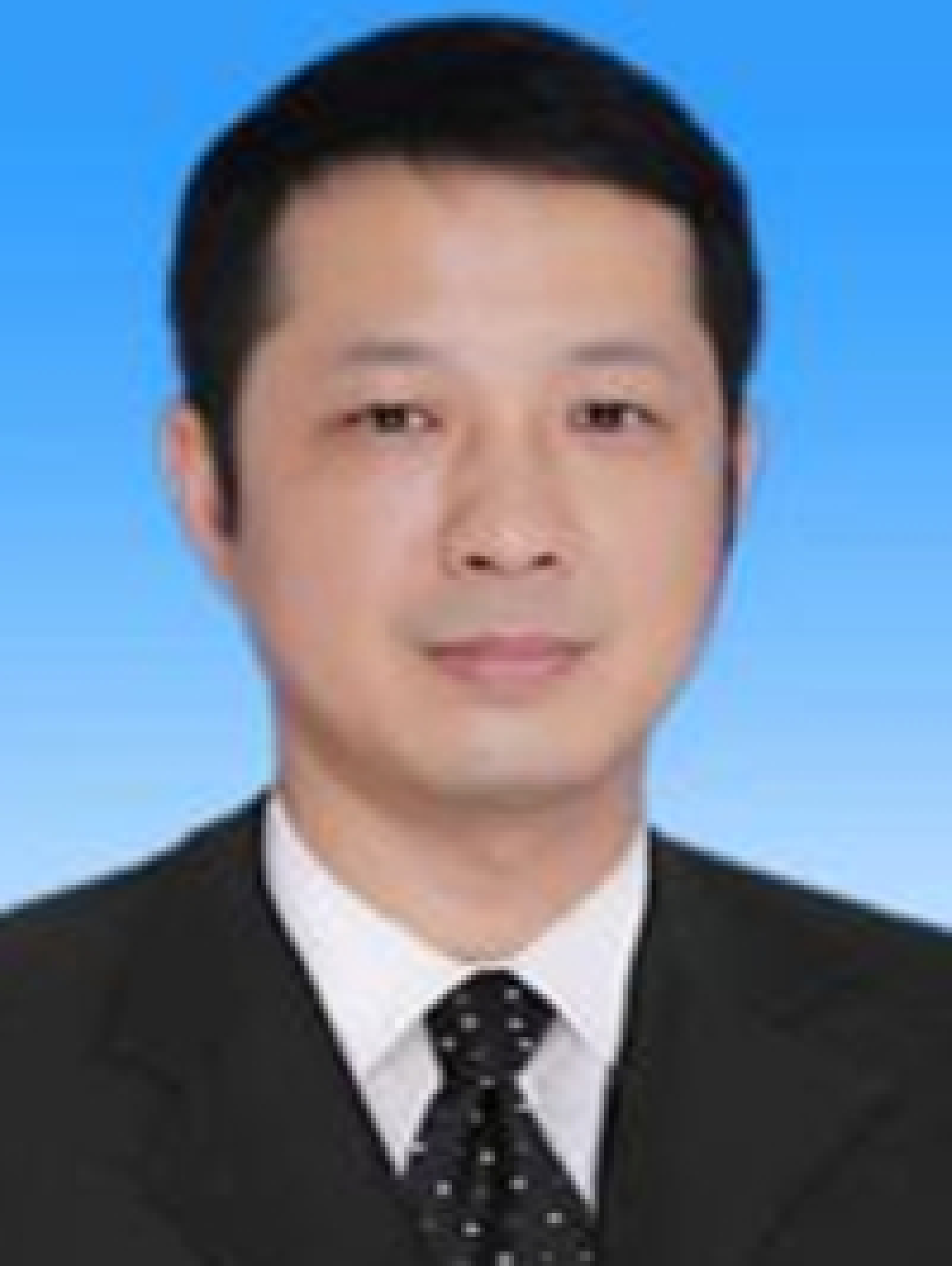}}]{Bing Liu} received the B.S., M.S., and Ph.D. degrees in 2002, 2005, and  2013, respectively, from the China University of Mining and Technology, Xuzhou, China. He is currently an Associate Professor at the School of Computer Science and Technology, China University of Mining and Technology, China. His current research interests include natural language processing, image understanding, and deep learning.
\end{IEEEbiography}

\begin{IEEEbiography}[{\includegraphics[width=1in,height=1.25in,clip,keepaspectratio]{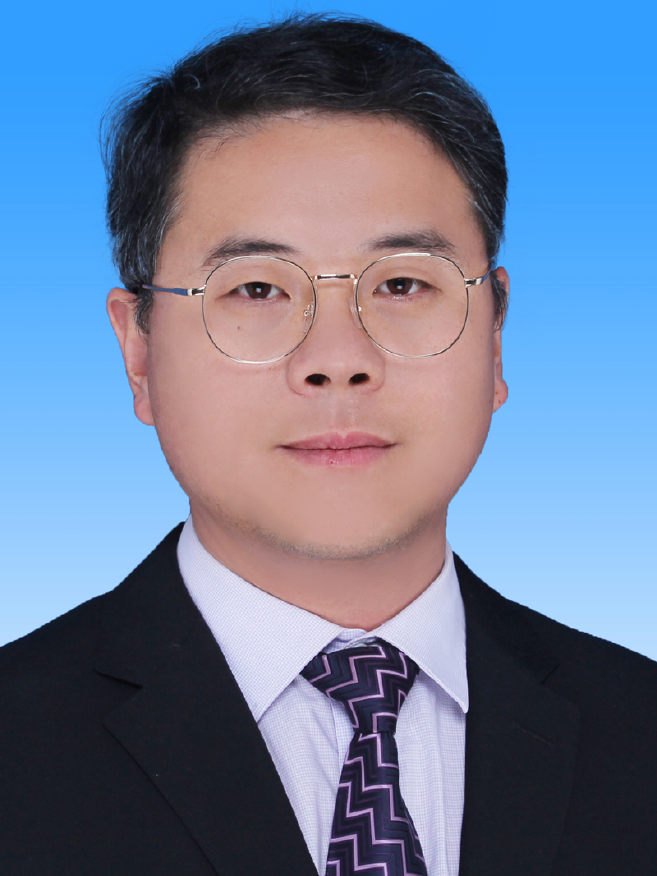}}]{Rui Yao} received the Ph.D. degree in computer science from the Northwestern Polytechnical University, Xi'an, China, in 2013. From 2011 to 2012, he was a Visiting Student with the University of Adelaide, Adelaide, SA, Australia. He is currently a Professor with the School of Computer Science and Technology, China University of Mining and Technology, Xuzhou, China. His research interests include computer vision and machine learning.
\end{IEEEbiography}

\end{document}